\documentclass[letterpaper]{article} 
\usepackage{aaai23}  
\usepackage{times}  
\usepackage{helvet}  
\usepackage{courier}  
\usepackage[hyphens]{url}  
\usepackage{graphicx} 
\usepackage{natbib}  
\usepackage{caption} 
\usepackage{algorithm}
\usepackage{algorithmic}

\usepackage{booktabs}
\usepackage{multirow}
\usepackage{graphicx}
\usepackage{svg}
\usepackage{amsmath,amssymb}
\usepackage{here}
\usepackage{url}
\usepackage{comment}
\usepackage{multirow}
\usepackage{tabularx}

\usepackage{newfloat}
\usepackage{listings}
\DeclareCaptionStyle{ruled}{labelfont=normalfont,labelsep=colon,strut=off} 
\floatstyle{ruled}
\newfloat{listing}{tb}{lst}{}
\floatname{listing}{Listing}
\newcommand{\EXiii}{$\mathcal{D}_{\rm Position} \cap \mathcal{D}_{\rm Word} \cap \mathcal{D}_{\rm Type}$~}

\newcommand{\EXioo}{$\mathcal{D}_{\rm Position} \cap \overline{\mathcal{D}_{\rm Word}} \cap \overline{\mathcal{D}_{\rm Type}}$~}

\newcommand{\EXoio}{$\overline{\mathcal{D}_{\rm Position}} \cap \mathcal{D}_{\rm Word} \cap \overline{\mathcal{D}_{\rm Type}}$~}

\newcommand{\EXooi}{$\overline{\mathcal{D}_{\rm Position}} \cap \overline{\mathcal{D}_{\rm Word}} \cap \mathcal{D}_{\rm Type}$~}

\newcommand{\EXooo}{$\overline{\mathcal{D}_{\rm Position}} \cap \overline{\mathcal{D}_{\rm Word}} \cap \overline{\mathcal{D}_{\rm Type}}$~}

\newcommand{\FigureWidth}{3.65cm}
\newcommand{\MCSurfaceWidth}{1.65cm}

\newcommand{\MCii}{$\mathcal{D}_{\mathrm{Top\text{-}1}} \cap \mathcal{D}_{\mathrm{Overlap}}$~}

\newcommand{\MCio}{$\mathcal{D}_{\mathrm{Top\text{-}1}} \cap \overline{\mathcal{D}_{\mathrm{Overlap}}}$~}

\newcommand{\MCoi}{$\overline{\mathcal{D}_{\mathrm{Top\text{-}1}}} \cap \mathcal{D}_{\mathrm{Overlap}}$~}

\title{Which Shortcut Solution Do Question Answering Models Prefer to Learn?}
\author{
    Kazutoshi Shinoda\textsuperscript{\rm 1,2},
    Saku Sugawara\textsuperscript{\rm 2},
    Akiko Aizawa\textsuperscript{\rm 1,2}
}
\affiliations{
    \textsuperscript{\rm 1}The University of Tokyo\\
    \textsuperscript{\rm 2}National Institute of Informatics\\
    shinoda@is.s.u-tokyo.ac.jp, \{saku, aizawa\}@nii.ac.jp
}

\begin{document}

\maketitle

\begin{abstract}
Question answering (QA) models for reading comprehension tend to learn shortcut solutions rather than the solutions intended by QA datasets.
QA models that have learned shortcut solutions can achieve human-level performance in shortcut examples where shortcuts are valid, but these same behaviors degrade generalization potential on anti-shortcut examples where shortcuts are invalid.
Various methods have been proposed to mitigate this problem, but they do not fully take the characteristics of shortcuts themselves into account.
We assume that the learnability of shortcuts, i.e., how easy it is to learn a shortcut, is useful to mitigate the problem.
Thus, we first examine the learnability of the representative shortcuts on extractive and multiple-choice QA datasets.
Behavioral tests using biased training sets reveal that shortcuts that exploit answer positions and word-label correlations are preferentially learned for extractive and multiple-choice QA, respectively.
We find that the more learnable a shortcut is, the flatter and deeper the loss landscape is around the shortcut solution in the parameter space.
We also find that the availability of the preferred shortcuts tends to make the task easier to perform from an information-theoretic viewpoint.
Lastly, we experimentally show that the learnability of shortcuts can be utilized to construct an effective QA training set; the more learnable a shortcut is, the smaller the proportion of anti-shortcut examples required to achieve comparable performance on shortcut and anti-shortcut examples.
We claim that the learnability of shortcuts should be considered when designing mitigation methods.
\end{abstract}

\section{Introduction}
Natural language understanding (NLU) models based on deep neural networks (DNNs) have been shown to exploit spurious correlations (also called dataset bias \cite{5995347} or annotation artifacts \cite{gururangan-etal-2018-annotation}) in the training set, and produce learning shortcut solutions \cite{geirhos_shortcut_2020} rather than the solutions intended by datasets.
Shortcut learning by NLU models causes poor generalization to anti-shortcut examples where the spurious correlations no longer hold and the learned shortcuts fail \cite{mccoy-etal-2019-right,gardner-etal-2020-evaluating}.

To date, question answering (QA) models for reading comprehension have been reported to learn several types of shortcut solutions \cite{jia-liang-2017-adversarial,sugawara-etal-2018-makes,ko-etal-2020-look}.
Various approaches have been proposed to mitigate these problems in QA, such as data augmentation \cite{shinoda-etal-2021-question} and debiasing methods \cite{ko-etal-2020-look,wu-etal-2020-improving}.
However, those methods have not fully taken the characteristics of shortcuts into account.

We assume that studying the learnability of each shortcut in QA datasets should be useful to construct training sets or design data augmentation methods for mitigating the problem.
This assumption is supported by the work by \citet{lovering2021predicting}, who show that the learnability of a shortcut and the proportion of anti-shortcut examples in a training set are the two important factors that affect the shortcut learning behavior in grammatical tasks.

To verify our assumption, we first examine the learnability of representative shortcuts in extractive and multiple-choice QA.
In addition, we investigate how the learnability of a shortcut is related to the proportion of anti-shortcut examples required to mitigate the shortcut learning.
Namely, we aim to answer the following research questions (RQs): \emph{1) When every shortcut is valid for answering every question in biased training sets, which shortcut do QA models prefer to learn? 2) Why are certain shortcuts learned in preference to other shortcuts from the biased training sets? 3) How quantitatively different is the learnability for each shortcut? 4) What proportion of anti-shortcut examples in a training set is required to avoid learning a shortcut? Is it related to the learnability of shortcuts?}

\begin{figure*}
    \centering
    \includegraphics[height=4.5cm,clip]{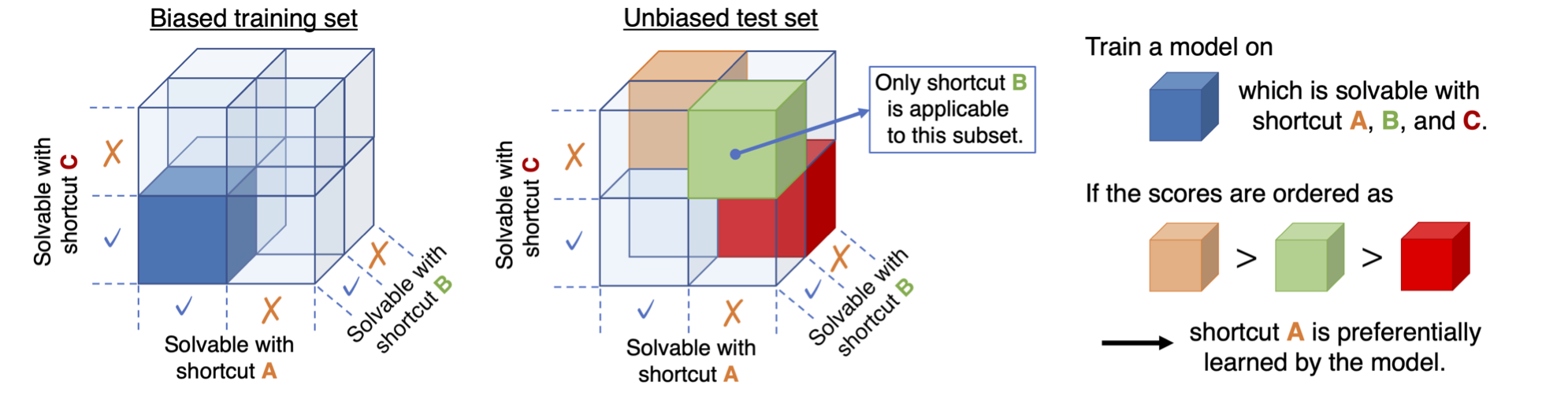}
    \caption{An illustration of the behavioral test to reveal which shortcut solution QA models prefer to learn.}
    \label{fig:illustration}
\end{figure*}

We answer the first question with behavioral tests using biased training sets as illustrated in Figure \ref{fig:illustration}.
These experiments reveal which shortcut solution is preferred by QA models when every shortcut is applicable to the biased training sets.
We show that, in extractive QA, the shortcut based on answer-position is preferred over the word matching and question-answer type matching shortcuts.
In multiple-choice QA, the shortcut exploiting word-label correlations is preferred to the one using lexical overlap.

We answer the second question from the perspective of the loss landscapes qualitatively.
We show that the flatness and depth of the loss surface around each shortcut solution in the parameter space can be the reason of the preference qualitatively.

To quantitatively explain the preference for shortcuts, we answer the third question by quantifying the learnability of shortcuts using the minimum description lengths.
We show that the availability of more preferred shortcuts in a dataset tend to make the task easier to learn.

Lastly, we answer the fourth question by simply changing the proportion of anti-shortcut examples in training sets and showing how the gap between the scores on shortcut and anti-shortcut examples changes.
We show that more learnable shortcuts require less proportion of anti-shortcut examples during training to achieve the comparable performance on shortcut and anti-shortcut examples.
Moreover, we find that only controlling the proportion of anti-shortcut examples is not sufficient to avoid learning less-learnable shortcuts.
Our findings suggest that the learnability of shortcuts should be considered when designing mitigation methods.

\section{Shortcut Solutions}

\subsection{Notation}

When a training or test set $\mathcal{D}$ of a dataset is given, we define a rule-based function for each shortcut ${\rm k}$ to split $\mathcal{D}$ into shortcut examples $\mathcal{D}_{\rm k}$ that are solvable with shortcut ${\rm k}$ and anti-shortcut examples $\overline{\mathcal{D}_{\rm k}}$ that are not solvable with shortcut ${\rm k}$.
Our rule-based functions are deterministic and easy to reproduce, while partial-input baselines that are widely used for detecting shortcut examples \cite{gururangan-etal-2018-annotation} depend on model choice and random seeds.

\subsection{Examined Shortcuts in Extractive QA}
\label{sec:examined-shortcut}
For extractive QA, we compared and analyzed the following three shortcuts, which were found in the existing literature.

\paragraph{Answer-Position}
Finding answers from the first sentence \cite{ko-etal-2020-look}: When QA models are trained on examples where answers are contained in the first sentence of the context, they learn to extract answers from the first sentence. (${\rm k}={\rm Position}$)

\paragraph{Word Matching}
Finding the answer from the most similar sentence \cite{sugawara-etal-2018-makes}: When an answer is contained in a sentence that is the most similar to a question, simple word matching is sufficient to find the correct answer.
We define the most similar sentence as the one that contains the longest n-gram in common with the question. (${\rm k}={\rm Word}$)

\paragraph{Type Matching}
Matching question and answer types \cite{weissenborn-etal-2017-making}: When the entity type of the answer to the question can be specified, and the textual spans corresponding to the expected answer type appear only once in the context, models can answer the question correctly by simply extracting the phrase of the entity type.
When the context contain two or more named entities of the same type as the answer, we classify the example into $\overline{\mathcal{D}_{\rm k}}$.
To define this shortcut rigorously, we omit answers that are not named entities.
We used spaCy \cite{spacy2} for named entity recognition.
(${\rm k}={\rm Type}$)

\begin{table}[tbp]
\centering
\begin{tabular}{lr|lr}
\toprule
\multicolumn{2}{c|}{RACE} & \multicolumn{2}{c}{ReClor} \\\midrule
\multicolumn{1}{c}{$w$} & \multicolumn{1}{c|}{$z^*$} & \multicolumn{1}{c}{$w$} & \multicolumn{1}{c}{$z^*$} \\\midrule
and & 23.6 & a & 6.7 \\
above & 20.7 & result & 5.3 \\
may & 20.7 & an & 5.1 \\
b & 16.5 & the & 4.9 \\
c & 13.5 & motive & 4.5\\
might & 10.5 & not & 4.3 \\
objective & 10.0 & stays & 4.2 \\
\bottomrule
\end{tabular}
\caption{Top 7 words with the highest z-statistics computed on RACE and ReClor training sets.}
\label{tab:biased-words-mcqa}
\end{table}

\begin{figure*}[t]
\centering
\small
\begin{tabular}{cccccc}
 & BERT & RoBERTa &  & BERT & RoBERTa \\
\rotatebox{90}{\hspace{2em}SQuAD 1.1} &
  \includegraphics[width=\FigureWidth]{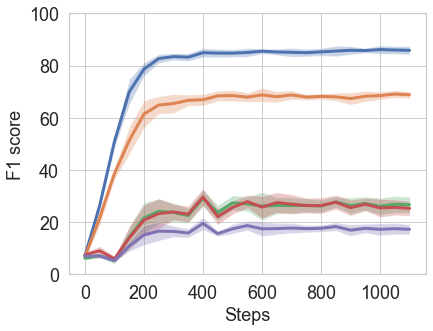} &
  \includegraphics[width=\FigureWidth]{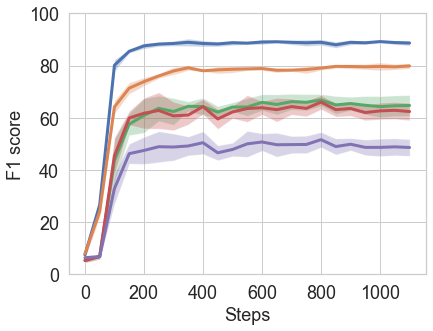} &
  \rotatebox{90}{\hspace{3em}RACE} & 
  \includegraphics[width=\FigureWidth]{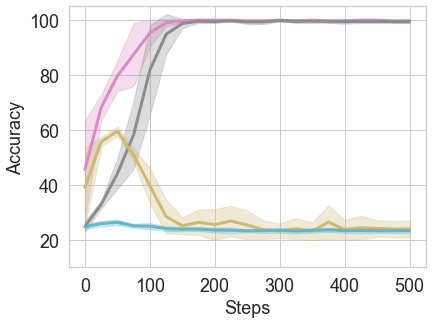} &
  \includegraphics[width=\FigureWidth]{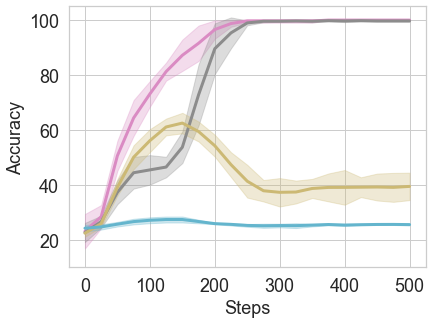} \\
\rotatebox{90}{\hspace{1em}NaturalQuestions}&
  \includegraphics[width=\FigureWidth]{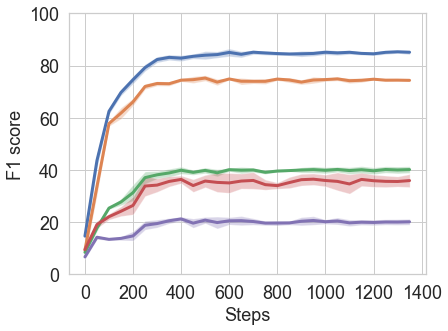} &
  \includegraphics[width=\FigureWidth]{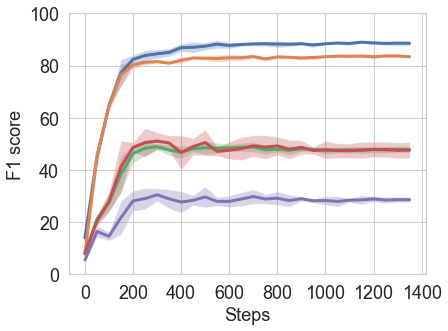} &
  \rotatebox{90}{\hspace{3em}ReClor} &
  \includegraphics[width=\FigureWidth]{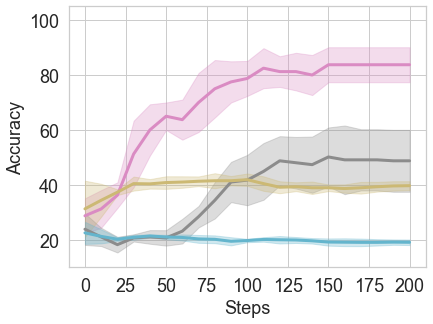} &
  \includegraphics[width=\FigureWidth]{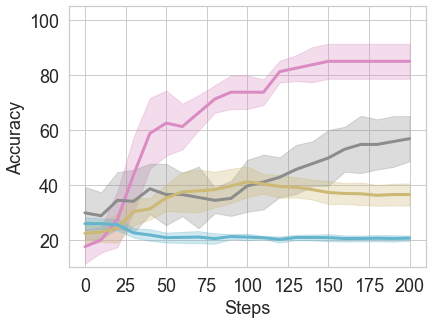} \\
\multicolumn{3}{c}{\includegraphics[height=2.8em]{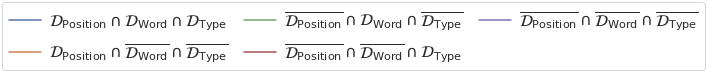}} & \multicolumn{3}{c}{\includegraphics[height=2.8em]{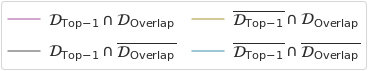}} \\
\end{tabular}
\caption{Left: F1 score on each subset of the SQuAD 1.1 and NaturalQuestions evaluation sets during training. Right: Accuracy on each subset of the RACE and ReClor test sets during training. The mean$\pm$standard deviations over 5 random seeds are displayed.}
\label{fig:training-dynamics}
\end{figure*}

\subsection{Examined Shortcuts in Multiple-choice QA}
For multiple-choice QA, we defined and analyzed the following two shortcuts.
We adopted the two shortcuts following the work on natural language inference (NLI) \cite{gururangan-etal-2018-annotation,mccoy-etal-2019-right} because multiple-choice QA and NLI are similar tasks as models predict whether the context+question (premise) entails the option (hypothesis).

\paragraph{Word-label Correlation}
Previous studies have shown that multiple-choice QA models can even make correct predictions with options only \cite{sugawara-etal-2020-assessing,yu2020reclor}.
NLI models can similarly make correct predictions with hypotheses only because certain words such as negation in hypotheses are highly correlated with labels \cite{gururangan-etal-2018-annotation}.
When considered in relation to the hypothesis-only bias in NLI, we assumed that multiple-choice QA datasets contain words in options that are highly correlated with binary labels.

Based on this assumption, we attempt to identify words in options that are highly correlated with the labels to define a realistic shortcut that exploits the word-label correlation.
\citet{gardner-etal-2021-competency} assumed that no single feature by itself should be informative about the class label.
Here, we generally follow their assumption.
We use z-statistics proposed by \citet{gardner-etal-2021-competency} to identify word $w$ in options with the conditional probability $p(y|w)$ that significantly deviates from the uniform distribution.
Specifically, we compute the z-statistics as
\begin{align}
z^* = \frac{p(y|w)}{\sqrt{p_0 (1 - p_0)/n}},
\end{align}
where $p_0$ is the uniform distribution of label $y$, $n$ is the frequency of word $w$, and $p(y|w)$ is the empirical distribution over $n$ samples where word $w$ is contained in the options.
$p_0$ is $1/4$ in RACE and ReClor datasets because they have four options for each question.
The top-7 words with the highest z-statistics in RACE and ReClor are shown in Table \ref{tab:biased-words-mcqa}.
We choose the top-1 word for the analysis of the word-label correlation shortcut for simplicity. (${\rm k}=\mathrm{Top\text{-}1}$)

\paragraph{Lexical Overlap}
NLI models exploit the lexical overlap between premise and hypothesis to make predictions \cite{mccoy-etal-2019-right}.
We assume that multiple-choice QA models can learn a similar shortcut solution using lexical overlap.
We define the lexical overlap shortcut as judging an option that has the maximum lexical overlap with context+question among the options to be the answer.
We define the lexical overlap as the ratio of the common uni-grams contained in both sequences to the number of words in an option. (${\rm k}=\mathrm{Overlap}$)

\section{Experiments}
\subsection{Experimental Setup}
\subsubsection{Datasets}
For extractive QA, we used SQuAD 1.1 \cite{rajpurkar-etal-2016-squad} and NaturalQuestions \cite{kwiatkowski-etal-2019-natural}, which contain more than thousand examples in the biased training sets in Figure \ref{fig:illustration}.
For multiple-choice QA, we used RACE \cite{lai-etal-2017-race} and ReClor \cite{yu2020reclor}, where option-only models can perform better than the random baselines \cite{sugawara-etal-2020-assessing,yu2020reclor}, suggesting that options in these datasets have unintended biases.

\subsubsection{Models}
We used BERT-base \cite{devlin-etal-2019-bert} and RoBERTa-base \cite{liu2019roberta} as encoders, which are widely adopted for extractive and multiple-choice QA \cite{yu2020reclor}.
The task-specific output layers were added on top of the encoders.
For extractive QA, models output the probability distributions of the start and end positions of answer spans over  context tokens.
For multiple-choice QA, models predicted the probability distribution of the correct option over four options.
The models were trained with cross-entropy loss minimization.
Except for the training steps, we followed the hyperparameters suggested by the original papers.\footnote{Our codes are publicly available at \url{https://github.com/KazutoshiShinoda/ShortcutLearnability}.}

\subsubsection{Evaluation Metrics}
For extractive QA, we used the F1 score as the evaluation metric, whereas for multiple-choice QA, we used accuracy.

\begin{figure*}[tbp]
\centering
\small
\begin{tabular}{c|c|c|c|c|c}
\toprule
\multicolumn{3}{c|}{SQuAD 1.1} & \multicolumn{3}{c}{NaturalQuestions} \\\midrule
Position & Word & Type & Position & Word & Type \\\midrule
\includegraphics[width=2cm]{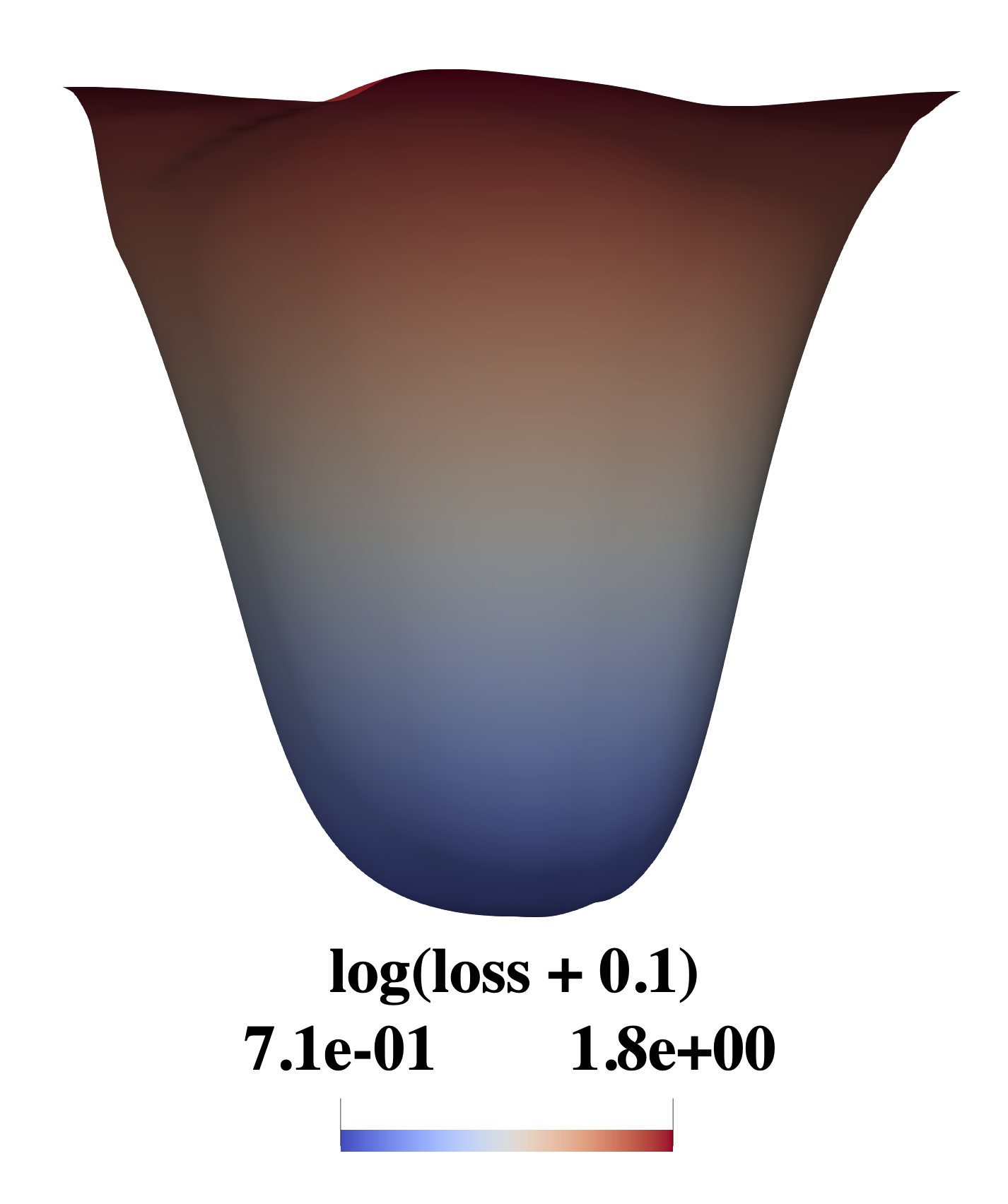} &
\includegraphics[width=2cm]{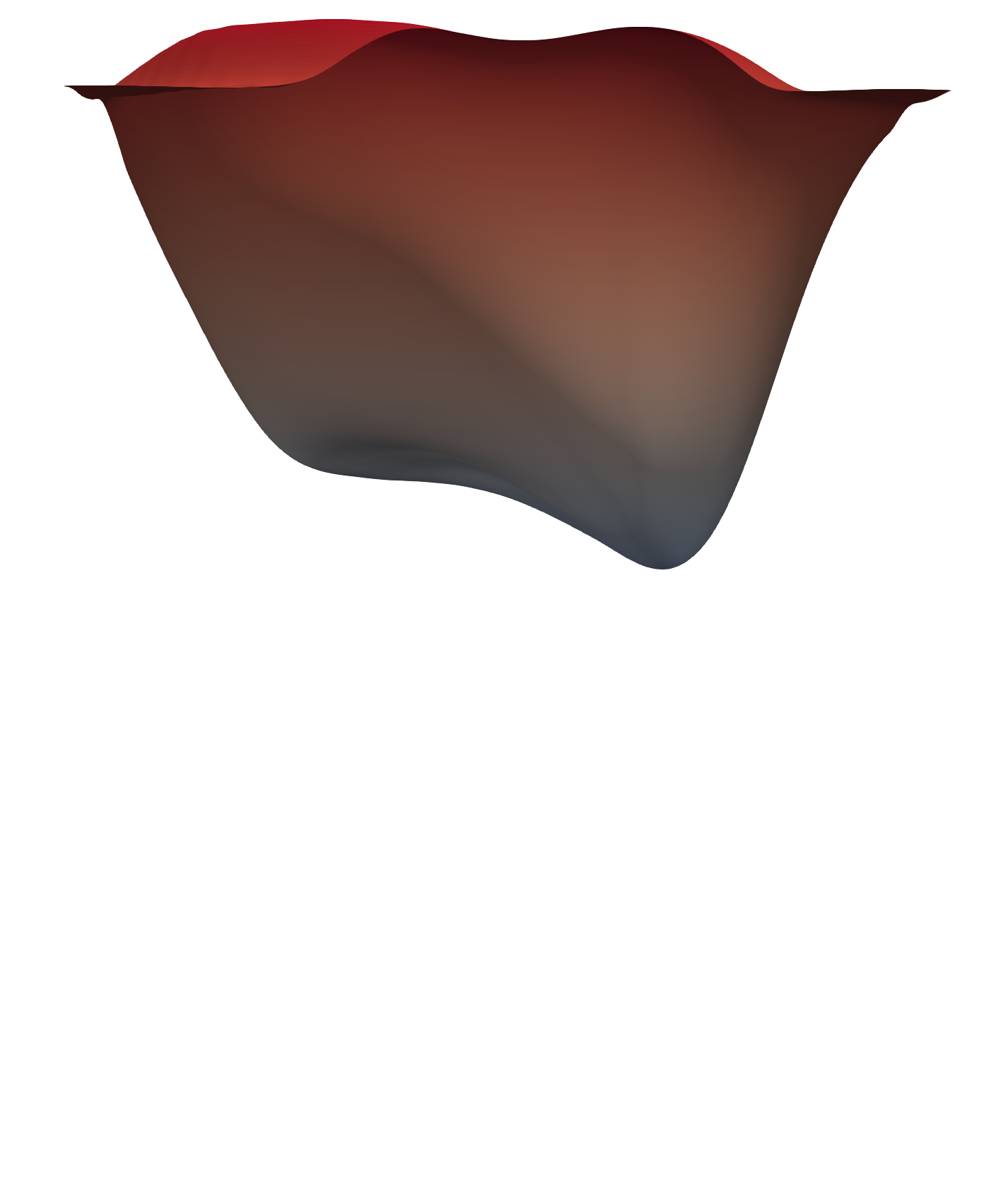} & \includegraphics[width=2cm]{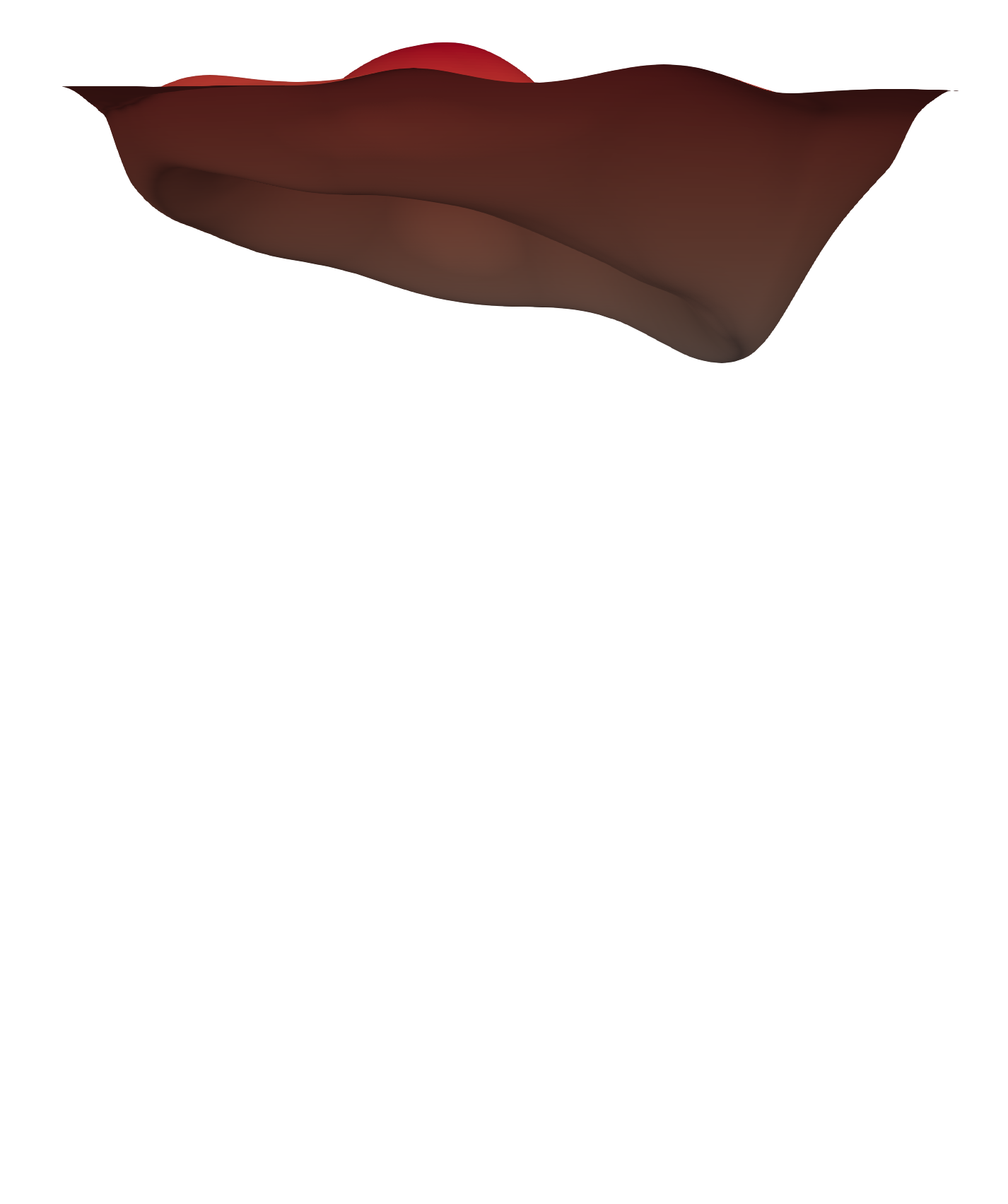} &
\includegraphics[width=2cm]{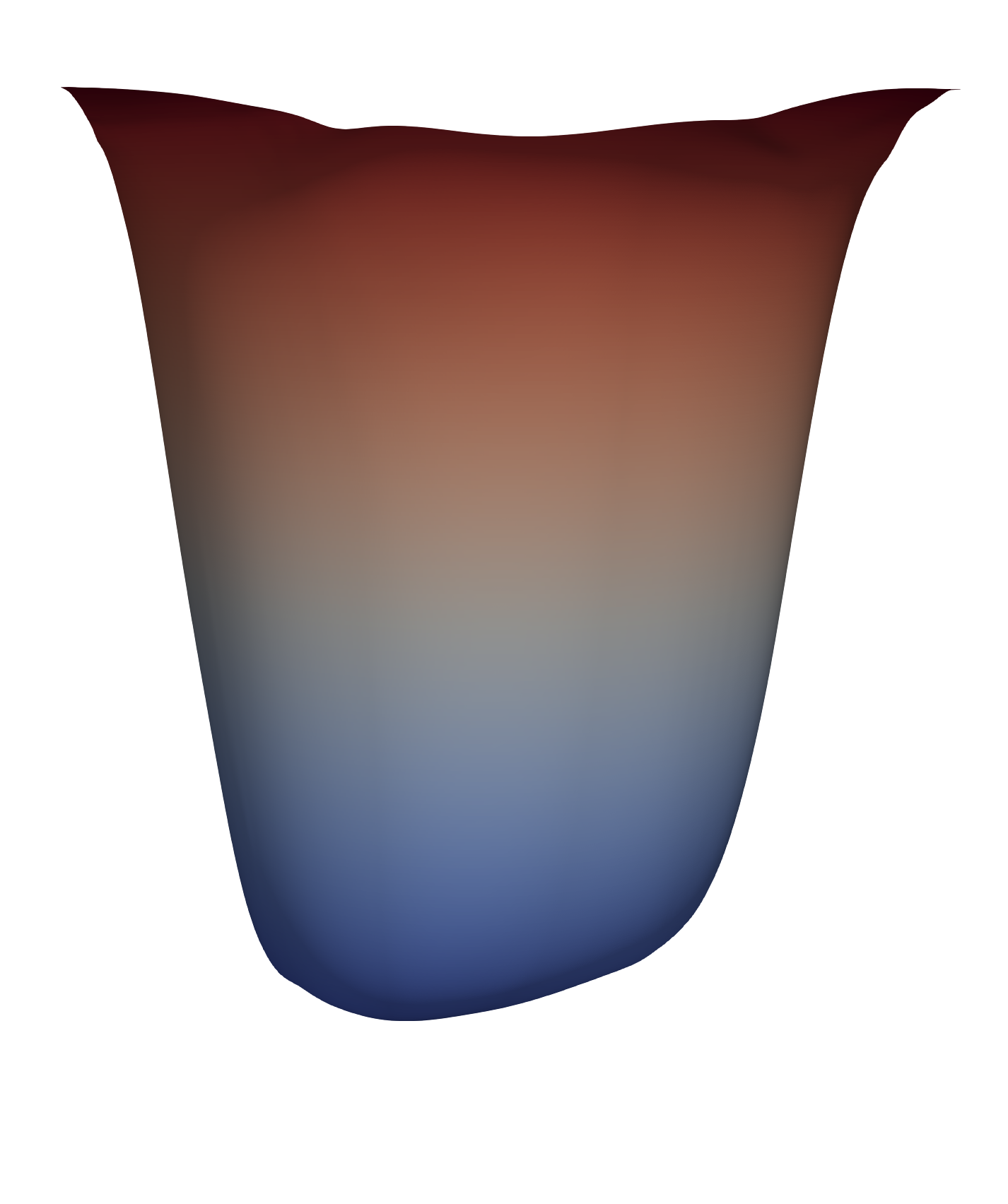} &
\includegraphics[width=2cm]{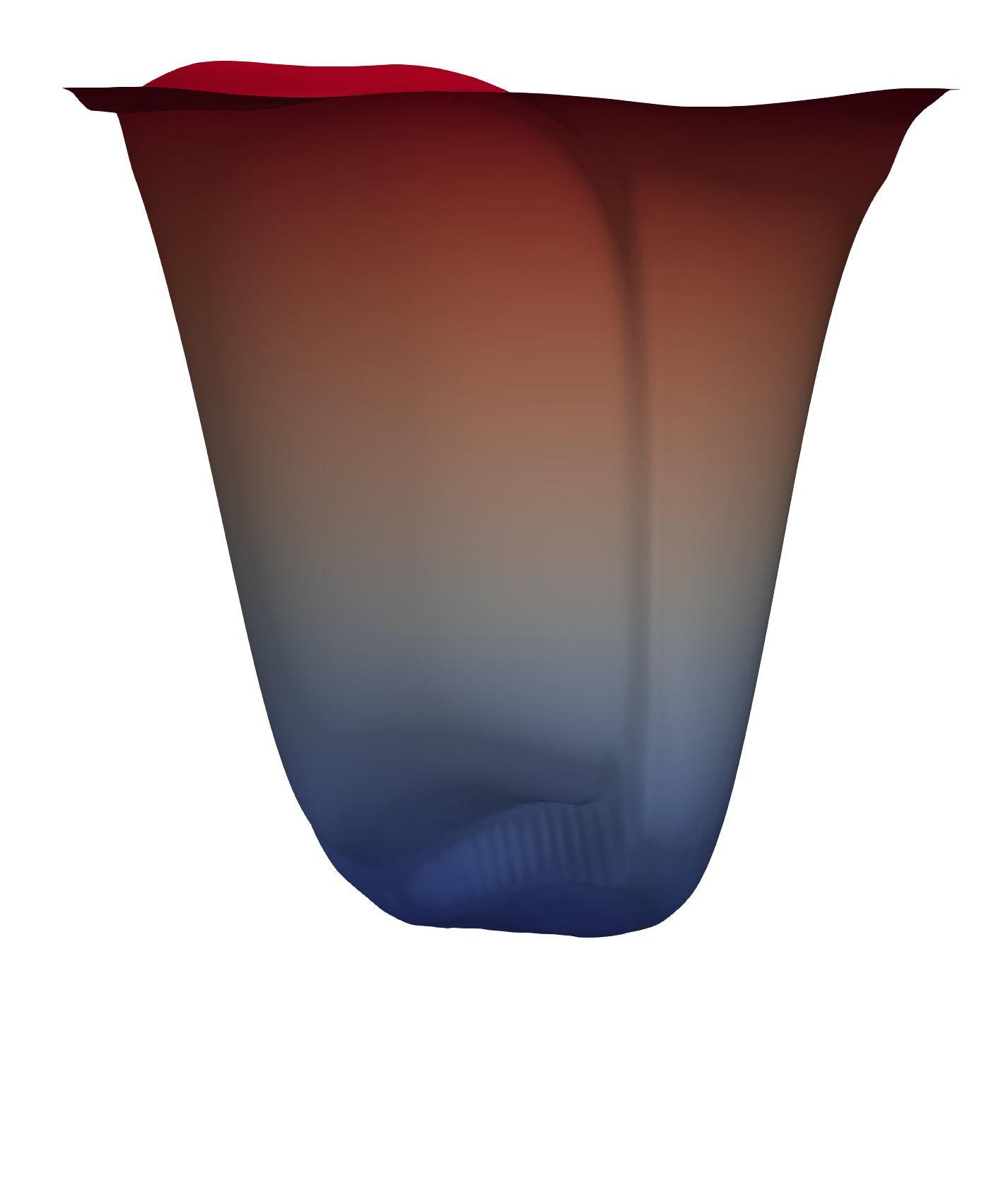} &
\includegraphics[width=2cm]{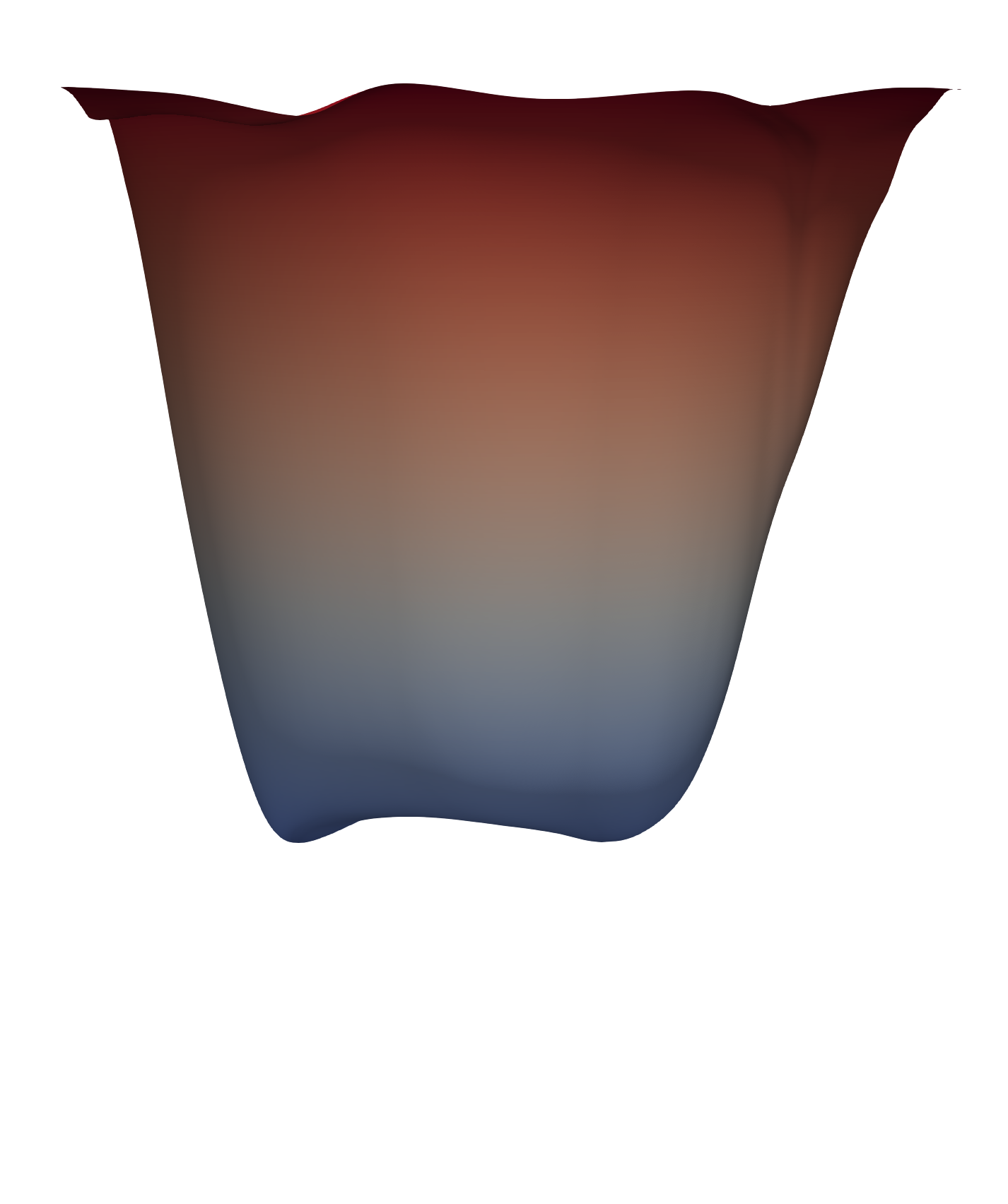} \\
\includegraphics[width=2cm]{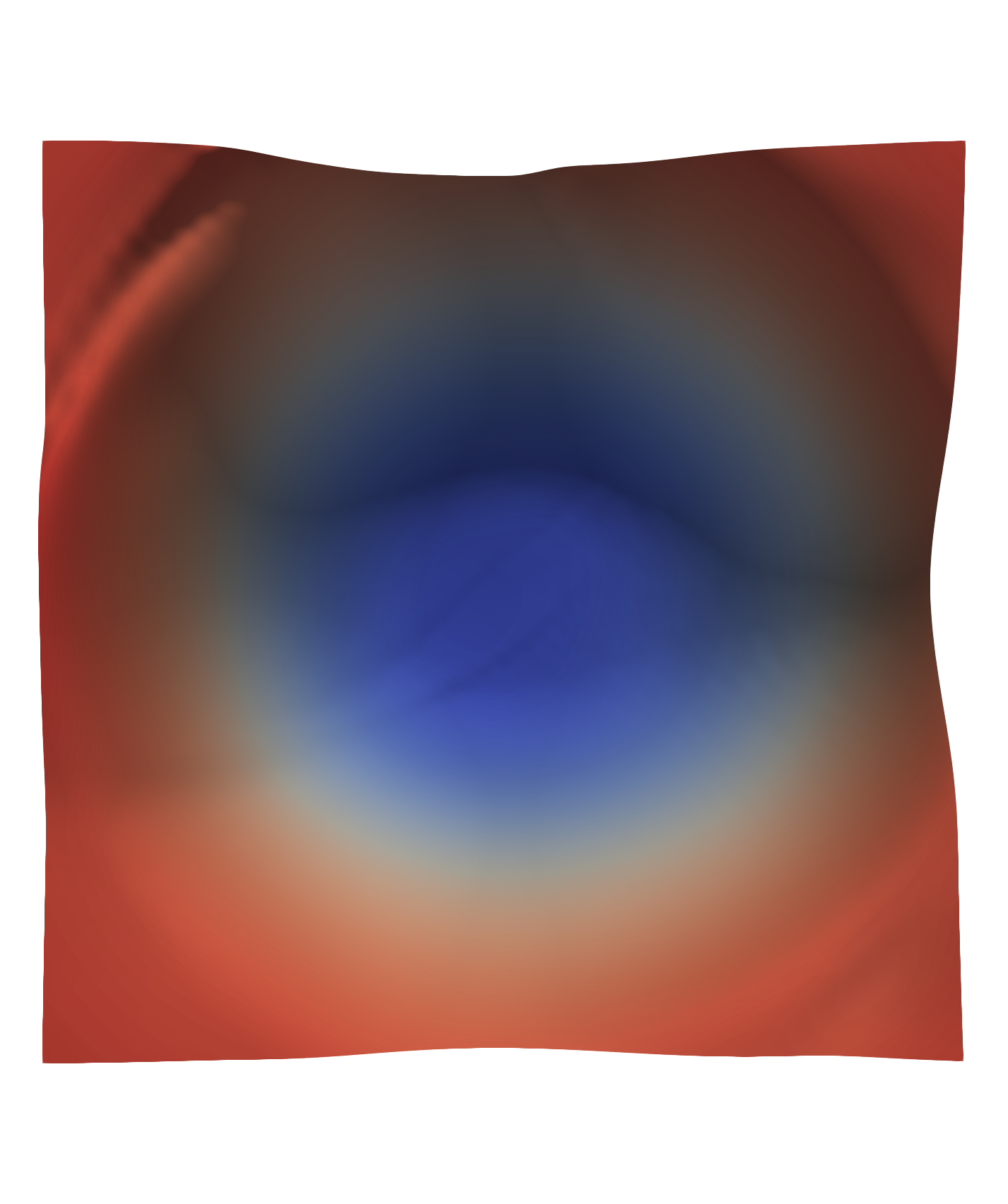} &
\includegraphics[width=2cm]{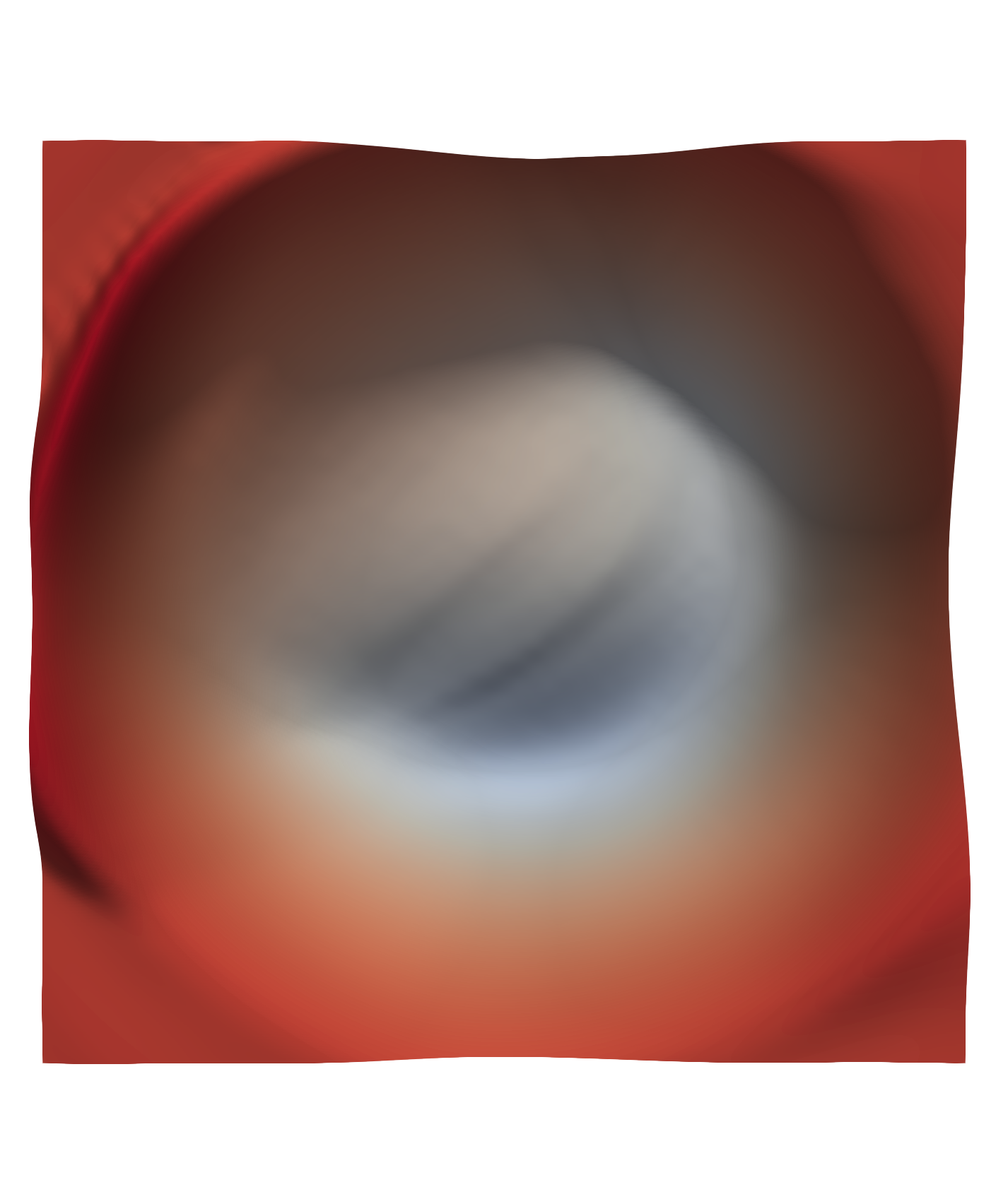} &
\includegraphics[width=2cm]{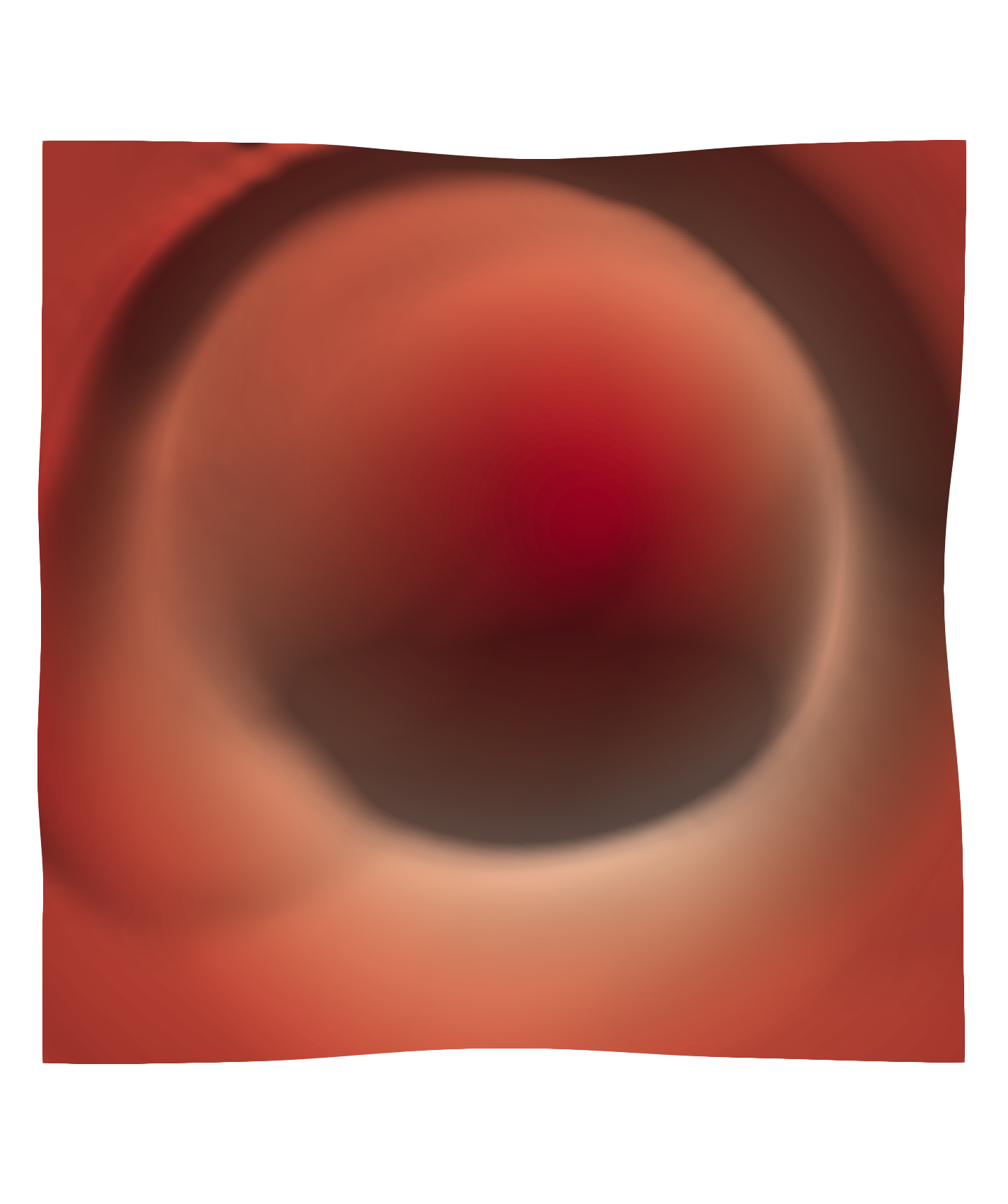} &
\includegraphics[width=2cm]{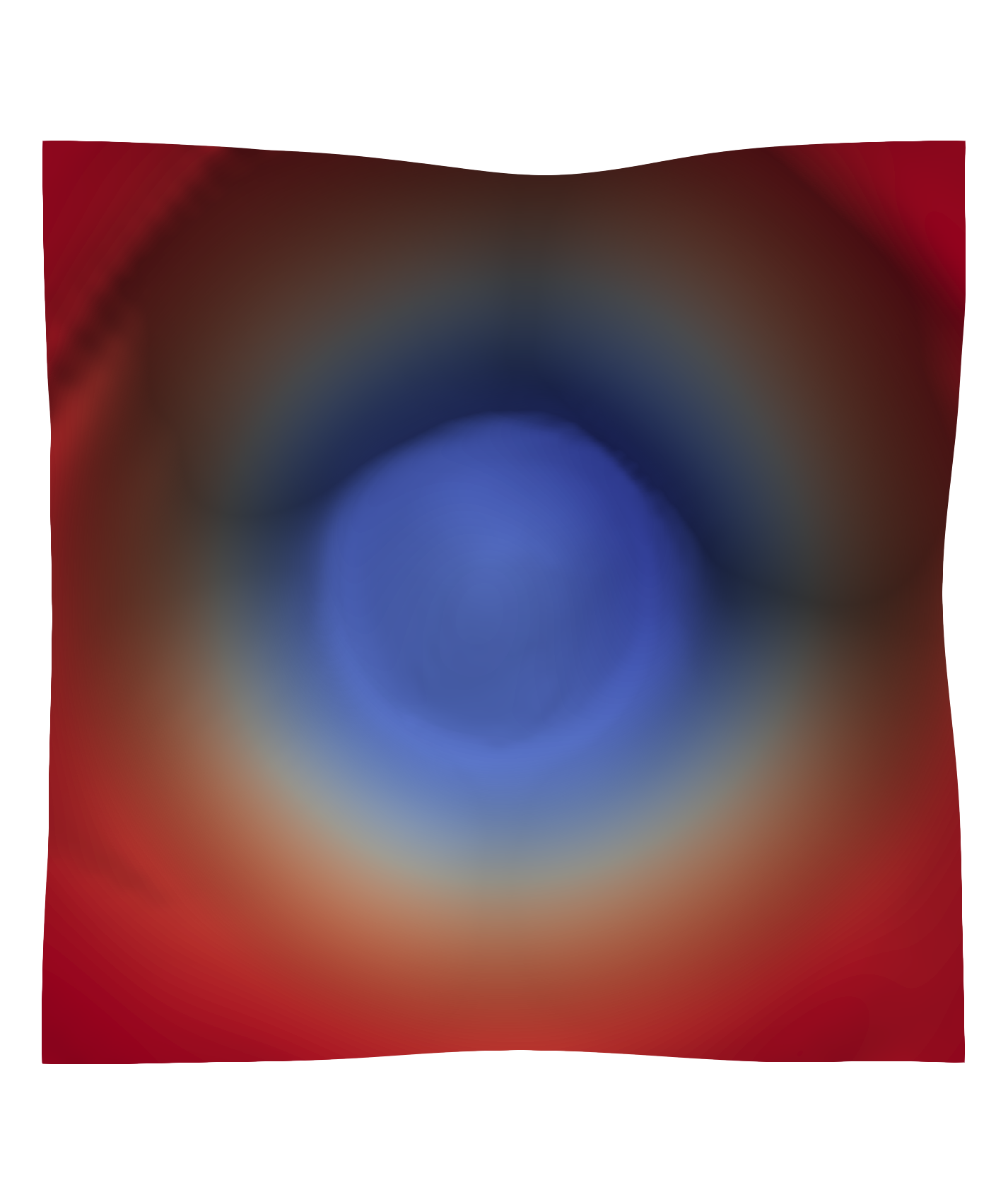} & \includegraphics[width=2cm]{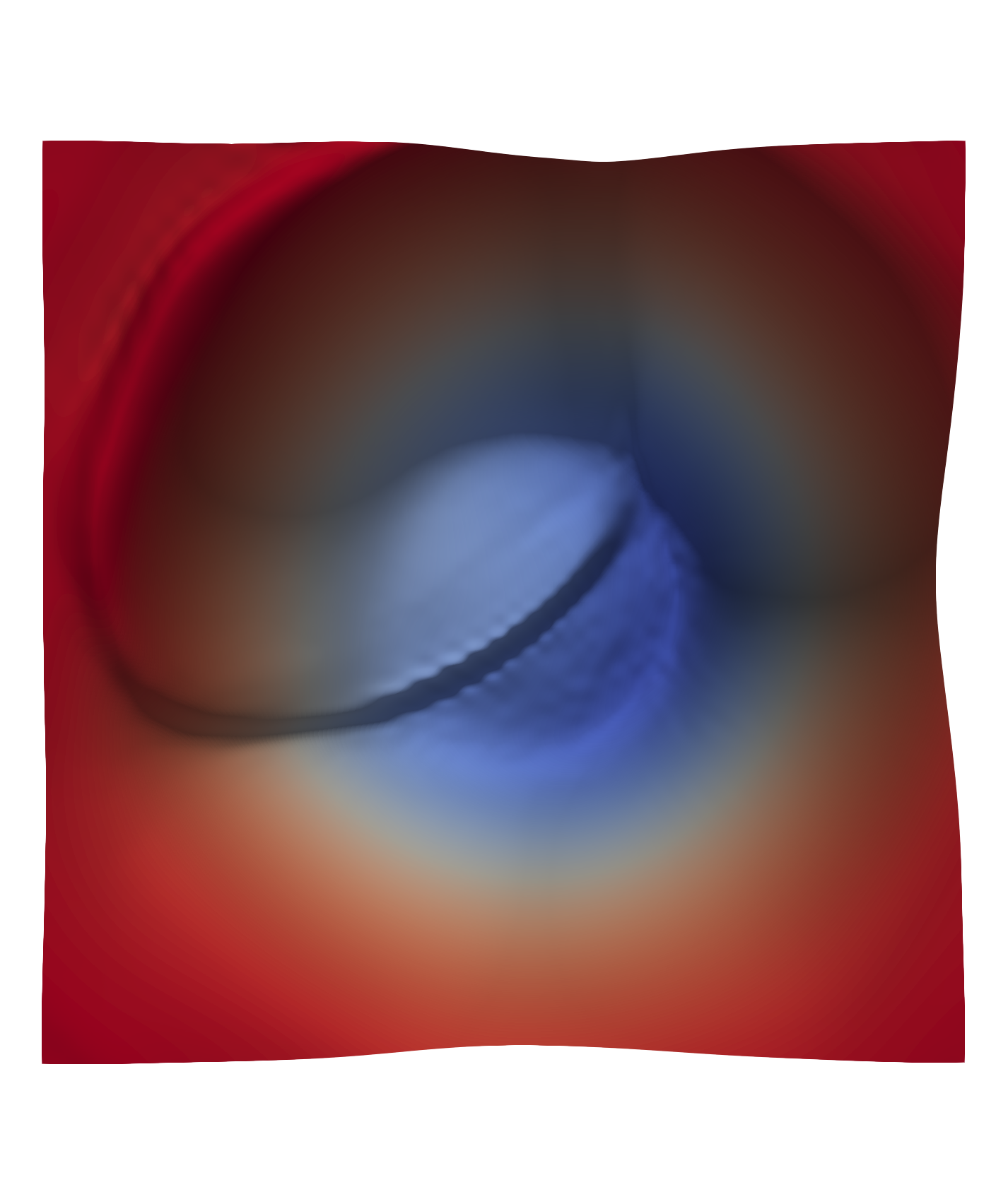} &
\includegraphics[width=2cm]{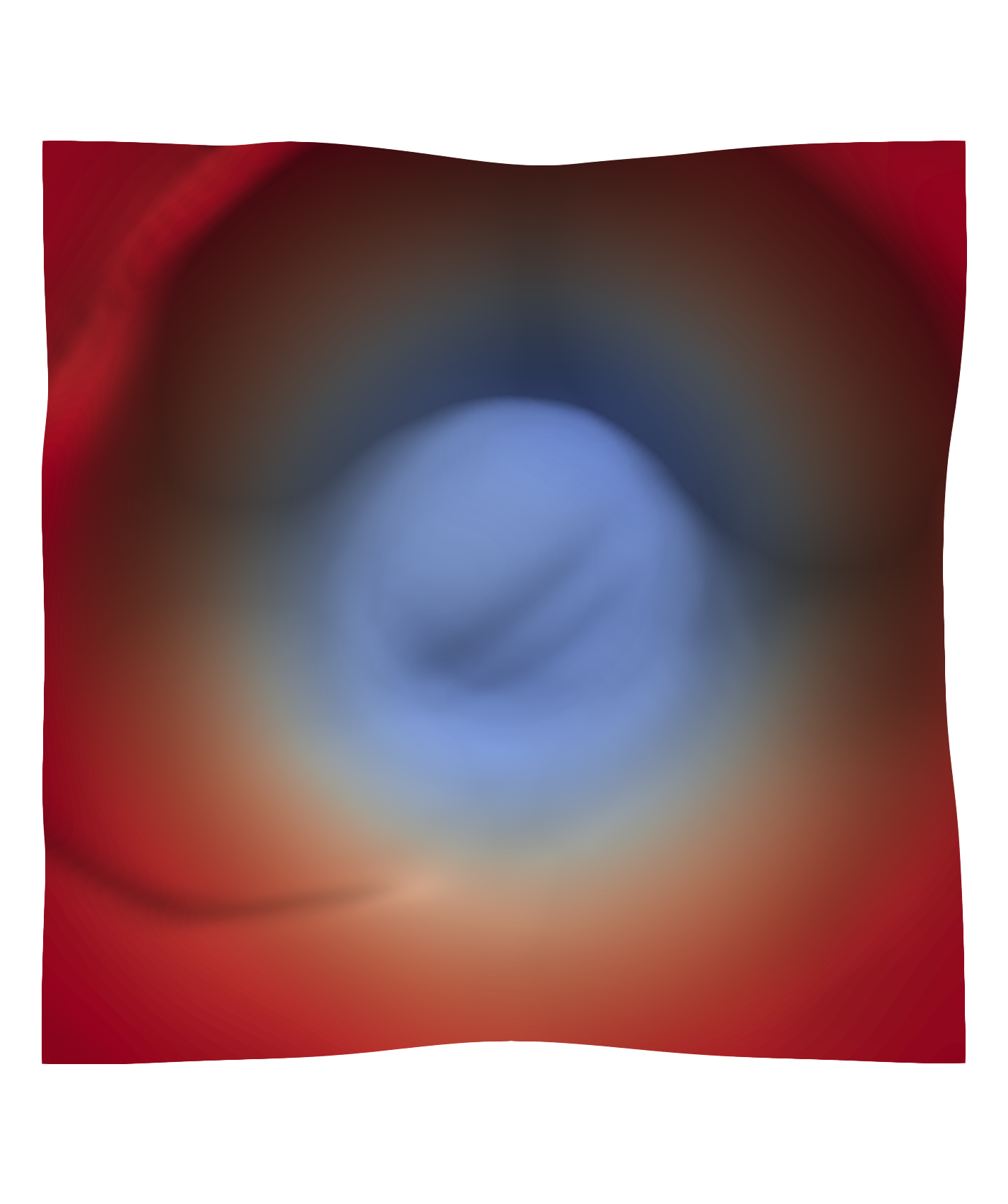} \\
\bottomrule
\end{tabular}
\caption{Visualization of loss landscapes around each shortcut in extractive QA datasets. The x and y directions are randomly selected in the parameter space. The center of the surface corresponds to the model that uses each shortcut.}
\label{fig:landscape-exqa}
\end{figure*}

\subsection{Learning from Biased Training Sets}
To compare the learnability of the examined shortcuts, we first answer the following research question (RQ).

\paragraph{RQ1} When every shortcut is valid for answering every question in biased training sets, which shortcut do QA models prefer to learn?

To answer this question, we conducted behavioral tests by training on a biased training set and testing on unbiased test sets as illustrated in Figure \ref{fig:illustration}.

The important factors of shortcut learning are 1) the frequency of anti-shortcut examples in a training set and 2) how easy it is to learn the shortcut from shortcut examples \cite{lovering2021predicting}.
In our biased training sets, all the examples are equally solvable with the examined shortcuts.
Therefore, our biased training enabled the impact of pure learnability to be compared.

\paragraph{Setup}
We first trained the models on \EXiii sampled from the training sets.
Then, the models were evaluated on subsets such as \EXioo sampled from the evaluation sets to clarify which shortcut models learn preferentially.
To gain insights into the process of learning shortcut solutions, we also examined the scores during training.

\paragraph{Results of Extractive QA}
Figure \ref{fig:training-dynamics} (left) shows the F1 score on each subset of the extractive QA datasets during training.
We assume that the higher the score on a subset where only one of the three shortcuts is valid, the more preferentially the model learns the shortcut.

Regardless of the datasets and models, the F1 score on \EXioo is higher than the F1 scores on \EXoio and \EXooi throughout the training.
This observation supports that, among the three, the shortcut using answer-position is the most learnable. 

Moreover, the scores on \EXioo increased significantly during the first several hundred training steps.
This observation is consistent with the experimental \cite{utama-etal-2020-towards,lai-etal-2021-machine} and theoretical results \cite{10.5555/3495724.3497160}; neural networks learn simpler functions at the early phase of training.

Conversely, the F1 scores on \EXoio and \EXooi were higher than that on \EXooo.
If the models exclusively learned the answer-position shortcut, the scores on these subsets would be similarly low regardless of the availability of the word and type matching shortcuts.
Therefore, this observation implies that the models did not exclusively learn only one shortcut, but a mixture of multiple shortcuts.

Of the two models, RoBERTa generalized better to \EXooo.
RoBERTa is able to learn sophisticated solutions other than the predefined shortcuts.
As BERT and RoBERTa have the same model architecture, the observations show that initialization points also affect the shortcut learning behavior.

\paragraph{Results of Multiple-choice QA}
Figure \ref{fig:training-dynamics} (right) shows the accuracy curve on each subset of the multiple-choice QA datasets during training.
At the end of the training, regardless of the models and the datasets, models learned to exploit word-label correlations more preferentially than lexical overlap because the accuracy on \MCio is ultimately greater than that on \MCoi at the end.

Interestingly, learning the shortcut using lexical overlap conversely took precedence over the shortcut using word-label only at the early stage of the training.
This may be because recognizing the dataset-specific word-label correlation requires hundreds of training steps as statistical evidence, while transformer-based language models might be originally equipped to recognize lexical overlap via self-attention \cite{NIPS2017_3f5ee243}.

\begin{figure}[tbp]
\centering
\small
\begin{tabular}{c|c|c|c}
\toprule
\multicolumn{2}{c|}{RACE} & \multicolumn{2}{c}{ReClor} \\\midrule
Top-1 & Overlap & Top-1 & Overlap \\
\midrule
\includegraphics[width=\MCSurfaceWidth]{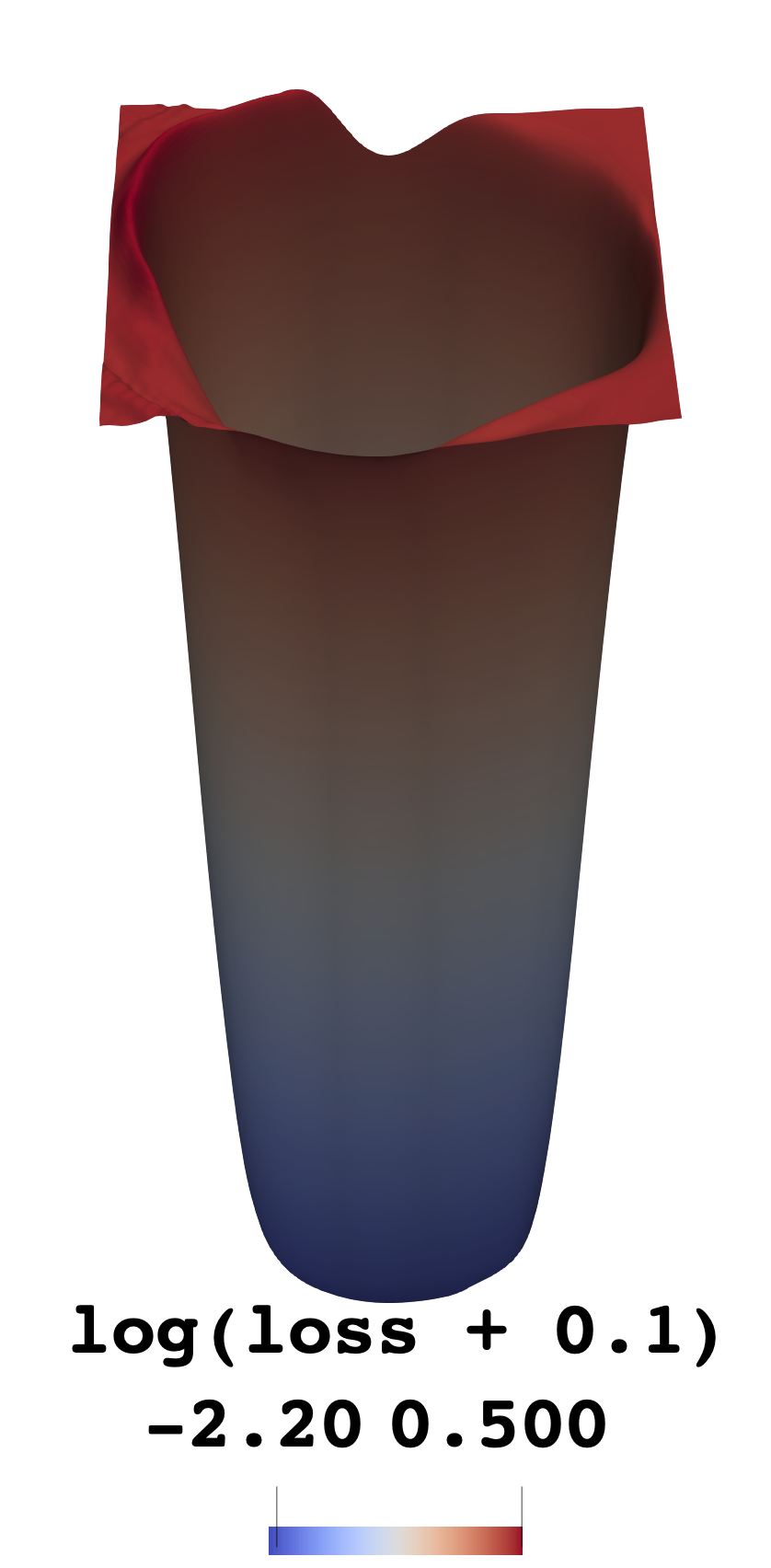} &
\includegraphics[width=\MCSurfaceWidth]{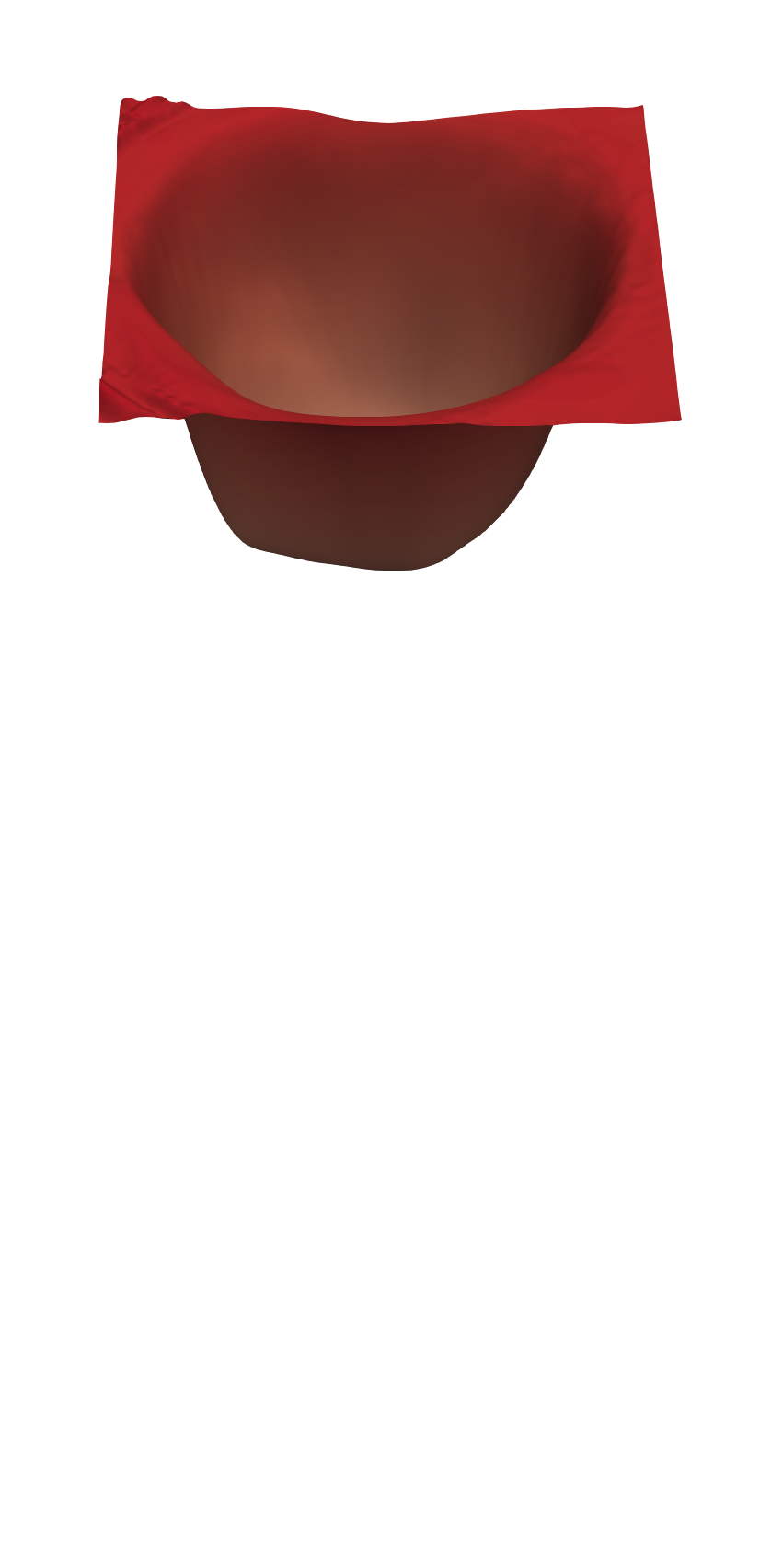} &
\includegraphics[width=\MCSurfaceWidth]{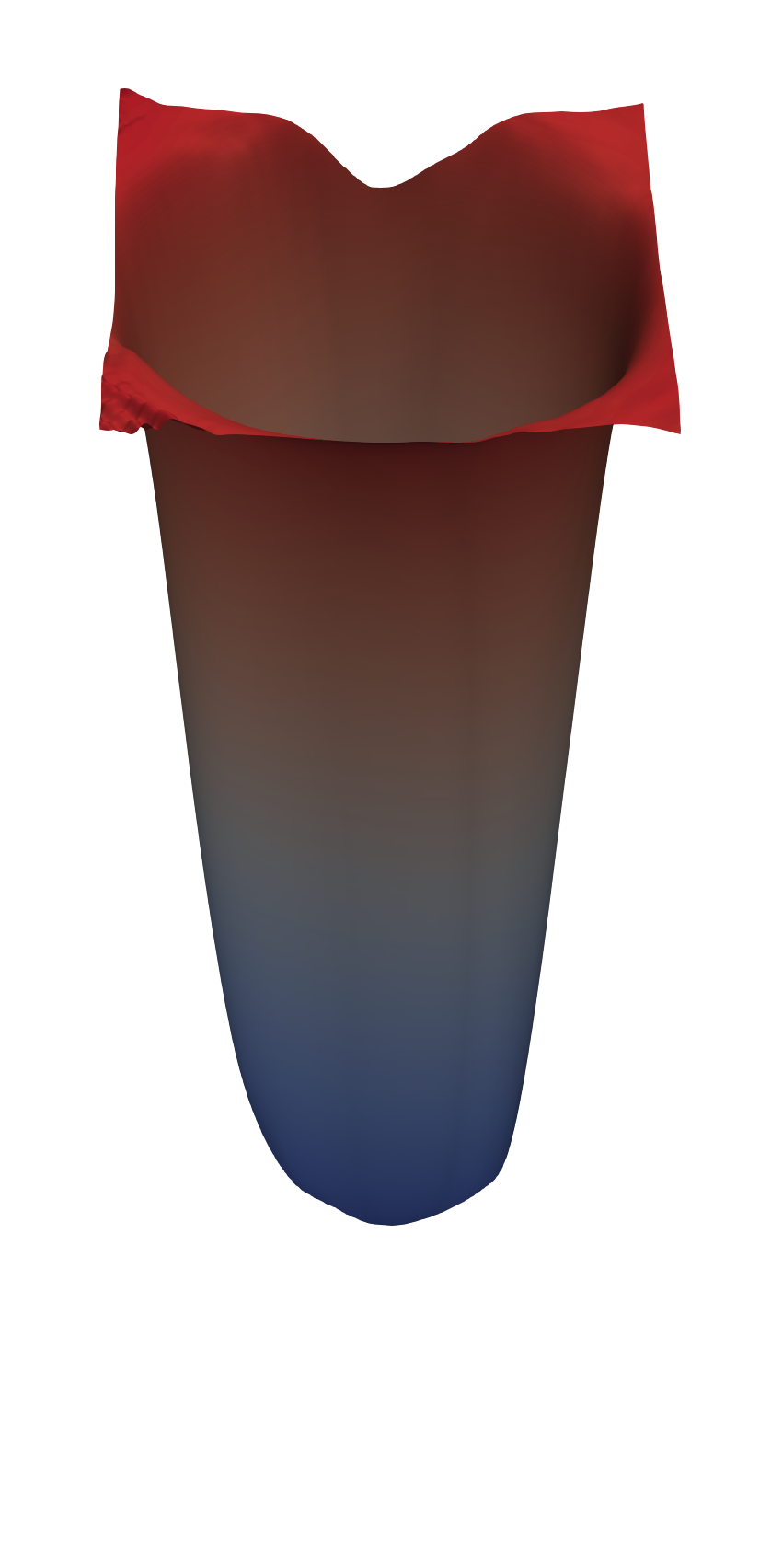} &
\includegraphics[width=\MCSurfaceWidth]{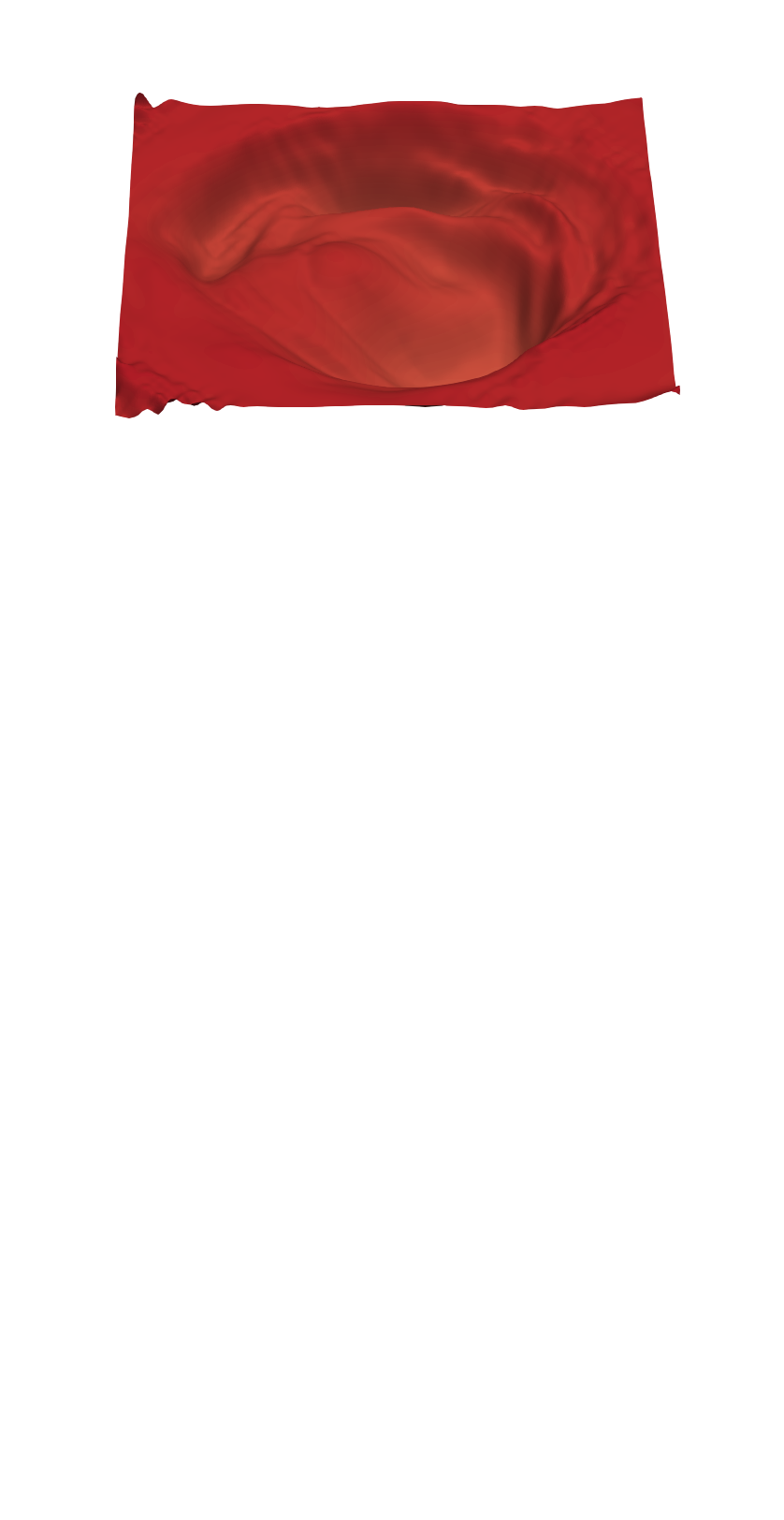} \\
\includegraphics[width=\MCSurfaceWidth]{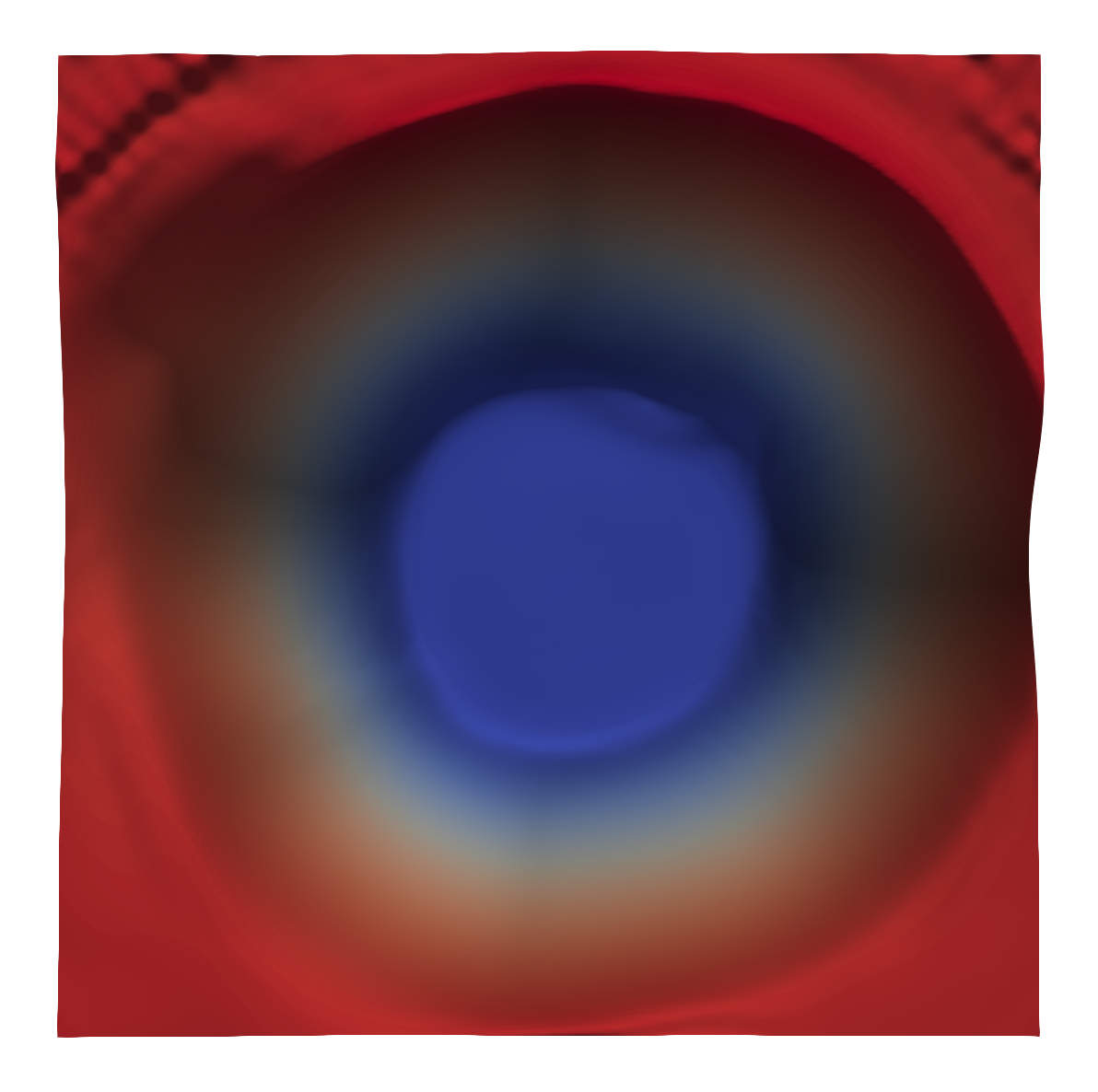} &
\includegraphics[width=\MCSurfaceWidth]{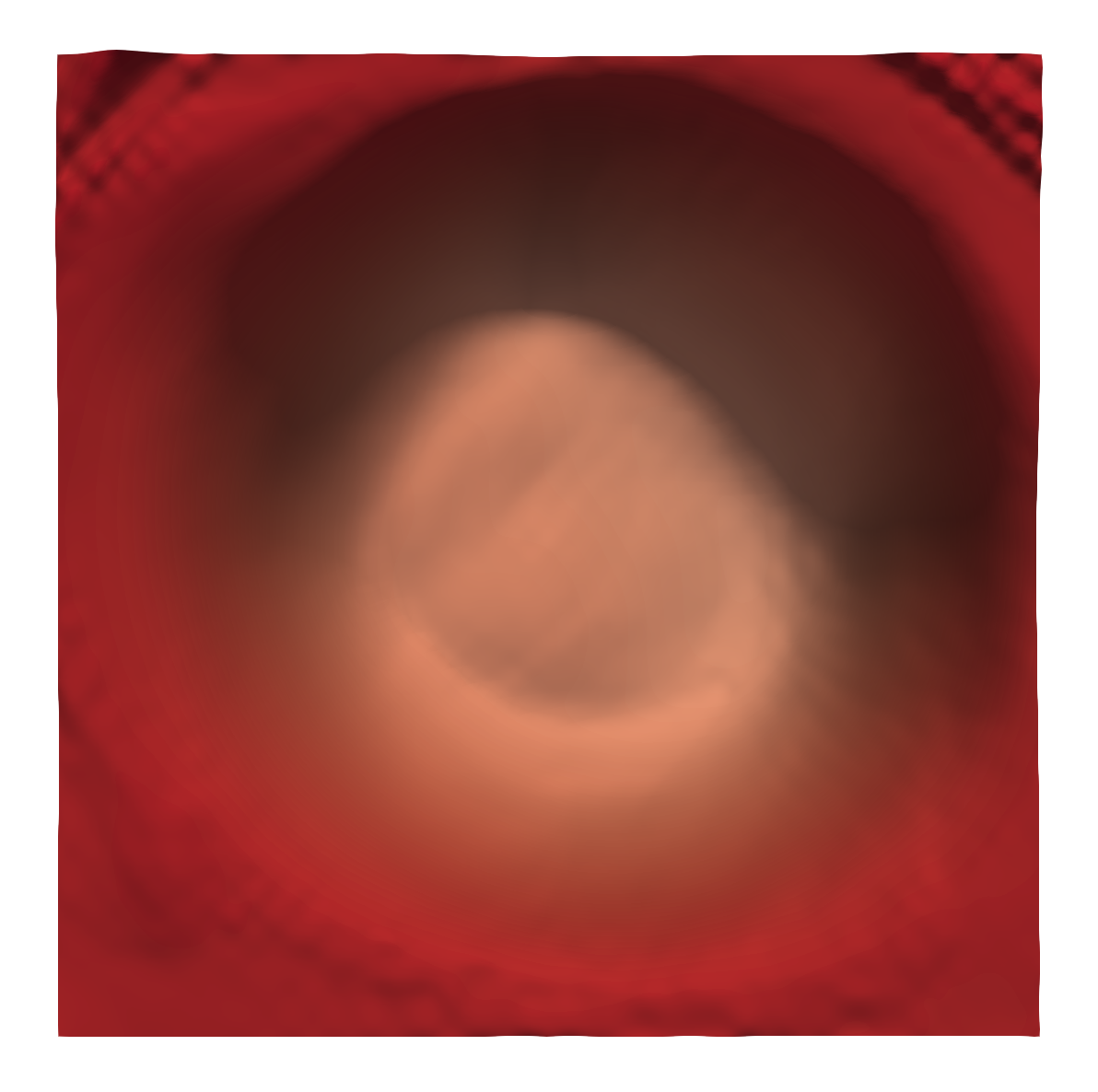} &
\includegraphics[width=\MCSurfaceWidth]{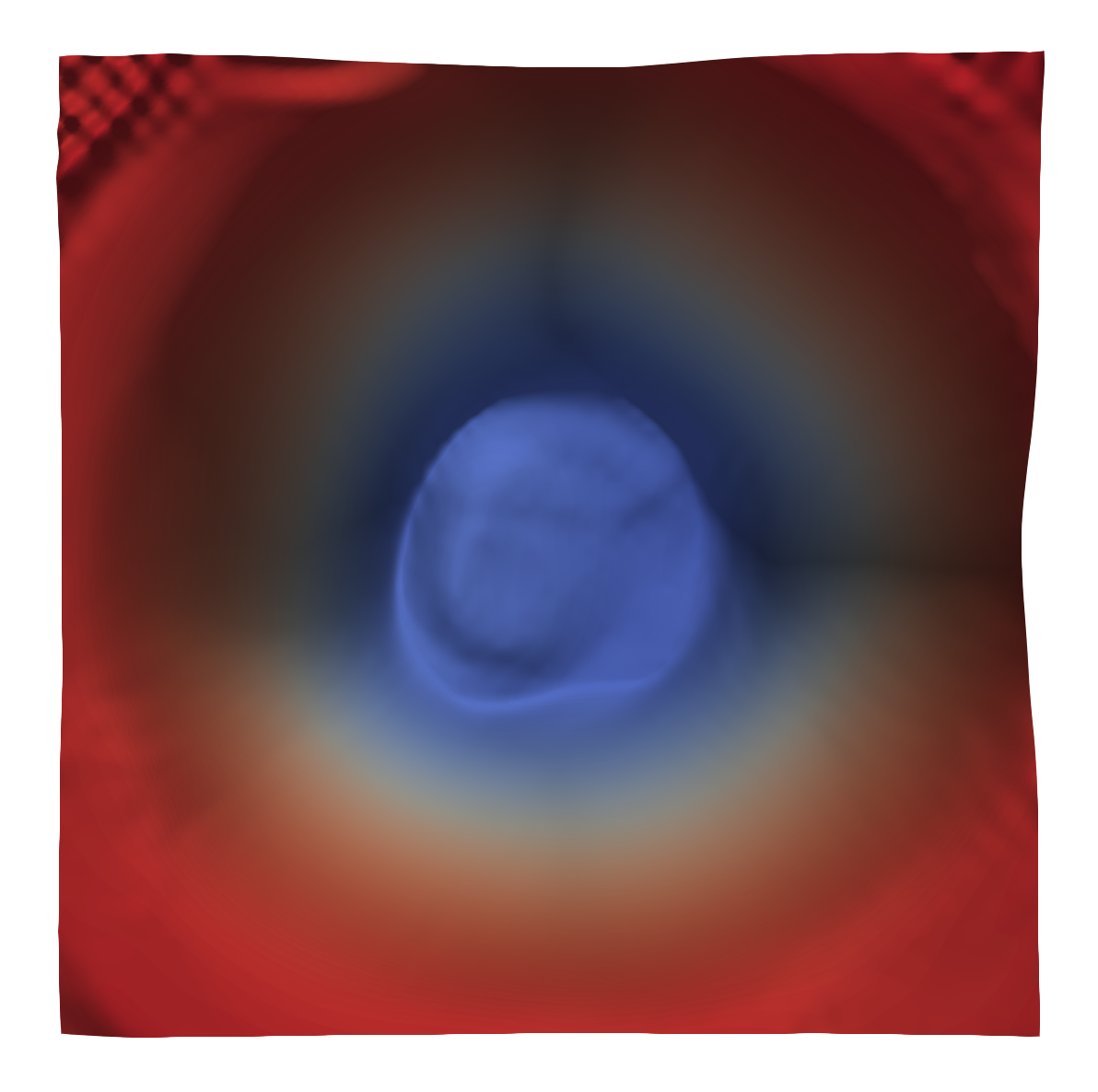} &
\includegraphics[width=\MCSurfaceWidth]{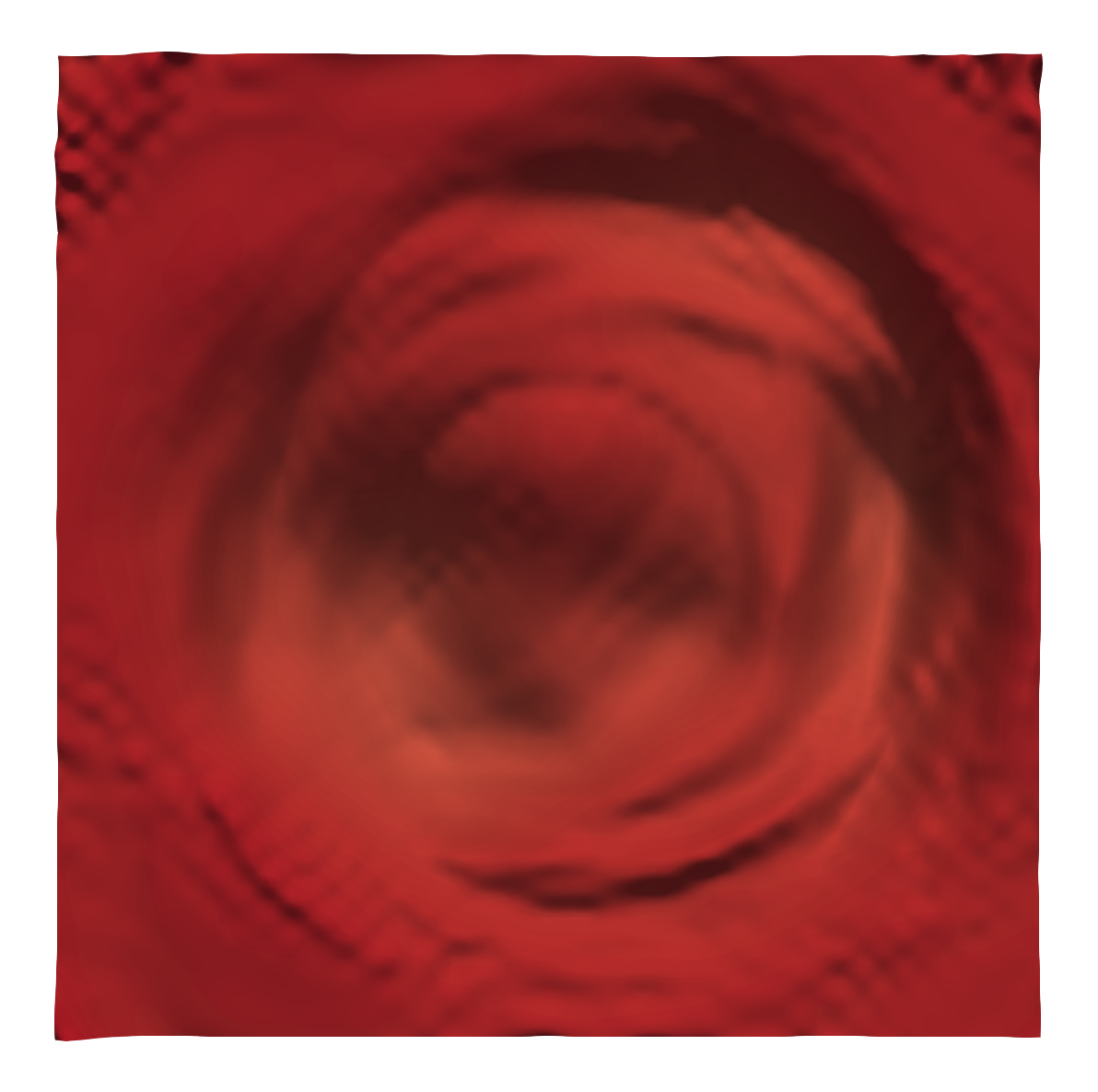} \\
\bottomrule
\end{tabular}
\caption{Visualization of loss landscapes around each shortcut in multiple-choice QA datasets.}
\label{fig:landscape-mcqa}
\end{figure}

\subsection{Visualizing the Loss Landscape}
\paragraph{RQ2} Why are certain shortcuts learned in preference to other shortcuts from the biased training sets?

We attempt to answer this question from the perspective of loss landscapes, as done by \citet{scimeca2022which} in image classification tasks.
Specifically, we visualize the loss landscapes around shortcut solutions and compare them.
The loss values were computed on subsets that are used as the biased training sets in the previous behavioral tests.
By doing so, we aim to compare the flatness of loss surfaces and gain insights into the preference.

\paragraph{Setup}
To visualize the loss landscape around a shortcut solution in the parameter space, we prepared models that use that shortcut.
We assume that models that are trained on subsets where only one shortcut is valid learn to use the shortcut.
For example, models trained on \EXioo are likely to exclusively learn the answer-position shortcut.
We verified this assumption by confirming that models achieved the best performance on the same subsets of the evaluation sets as the training sets.

For visualization, we first randomly selected two directions in the parameter space.
We displayed the loss values computed on \EXiii and \MCii on the hyperplane spanned by the two directions following \citet{NEURIPS2018_a41b3bb3}.

\paragraph{Results}
The visualization results for extractive and multiple-choice QA are displayed in Figures \ref{fig:landscape-exqa} and \ref{fig:landscape-mcqa}.
The center of each figure represents each shortcut solution.

The results show that the QA models that learn the preferred shortcuts (Position and Top-1) tend to lie in flatter and deeper loss surfaces.\footnote{We follow the definition of the flatness as the size of the connected region in the parameter space where the loss remains approximately constant \cite{10.1162/neco.1997.9.1.1}.}
The orders of the flatness and depth of the loss surfaces are roughly correlated with the preferential order of learning shortcuts in the previous behavioral tests.
These observations explain why models trained on \EXiii and \MCii learned to use the answer-position and word-label correlation shortcuts, respectively.

\subsection{Rissanen Shortcut Analysis}
\paragraph{RQ3} How quantitatively different is the learnability for each shortcut?

By answering this question, we aim to quantitatively explain the preference for shortcuts.
To this end, we approximately computed the minimum description length (MDL) \cite{RISSANEN1978465} on the biased datasets where one of the predefined shortcuts is applicable, such as \EXioo, and investigated how MDL changed for each shortcut.
Formally, MDL measures the number of bits needed to communicate the labels $y$ given the inputs $x$ in a biased subset of a dataset.
We name this method Rissanen Shortcut Analysis (RSA), after the father of the MDL principle.
Intuitively, RSA is simple yet effective to examine how well the availability of a shortcut in a training set makes the task easier to learn in a theoretically grounded manner.

\begin{table}[tbp]
\centering
\begin{tabular}{lrr}
\toprule
Shortcut & \multicolumn{1}{c}{BERT} & \multicolumn{1}{c}{RoBERTa} \\
\midrule
\multicolumn{3}{l}{\emph{SQuAD 1.1}} \\
~~~Position & 4.65 ± 0.12 & 4.22 ± 0.23 \\
~~~Word & 4.94 ± 0.24 & 3.73 ± 0.17 \\
~~~Type & 5.75 ± 0.30 & 4.52 ± 0.06 \\\midrule
\multicolumn{3}{l}{\emph{NaturalQuestions}}\\
~~~Position & 6.28 ± 0.15 & 5.37 ± 0.24 \\
~~~Word & 12.24 ± 0.14 & 9.08 ± 0.20 \\
~~~Type & 11.76 ± 0.55 & 8.83 ± 0.38 \\\midrule
\multicolumn{3}{l}{\emph{RACE}}\\
~~~Top-1 & 0.52 ± 0.34 & 0.41 ± 0.29 \\
~~~Overlap & 4.16 ± 0.55 & 3.55 ± 0.10 \\\midrule
\multicolumn{3}{l}{\emph{ReClor}}\\
~~~Top-1 & 0.33 ± 0.07 & 0.28 ± 0.03 \\
~~~Overlap & 0.55 ± 0.03 & 0.52 ± 0.02 \\\bottomrule
\end{tabular}
\caption{Minimum description lengths (kbits) on biased datasets where only one of the examined shortcut solutions is valid. The means$\pm$standard deviations over five random seeds are reported.}
\label{tab:mdl}
\end{table}

\begin{figure*}[tbp]
\centering
\begin{tabular}{ccc}
Position & Word & Type \\
\includegraphics[width=5cm]{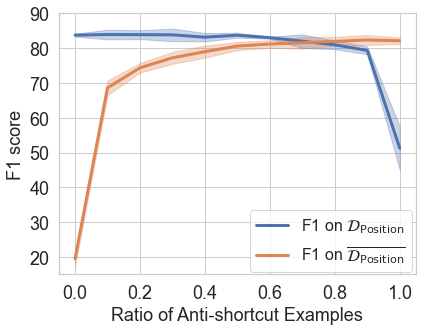} & \includegraphics[width=5cm]{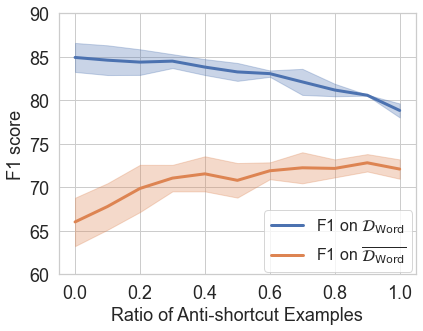} & \includegraphics[width=5cm]{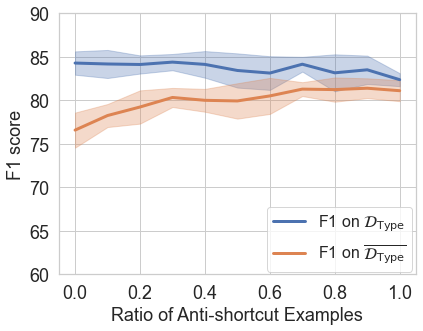} \\
\end{tabular}
\caption{F1 scores on shortcut and anti-shortcut examples from SQuAD with different proportions of anti-shortcut examples in the training set, with the size set to 5k. The mean$\pm$standard deviations over 5 random seeds are displayed.}
\label{fig:blend-anti-exqa}
\end{figure*}

\begin{figure*}[tbp]
\centering
\begin{tabular}{cc}
Top-1 & Overlap \\
\includegraphics[width=5cm]{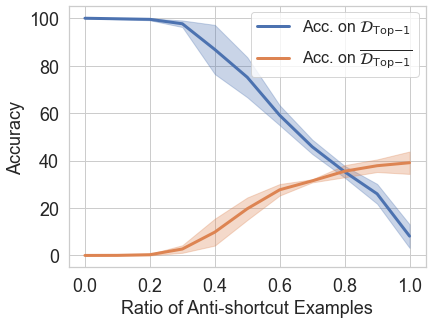} & \includegraphics[width=5cm]{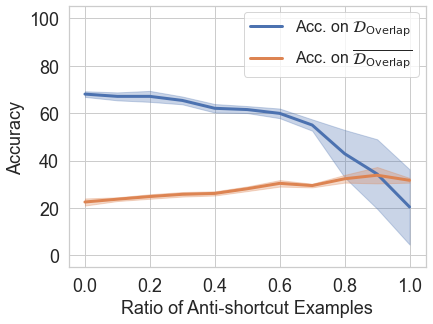} \\
\end{tabular}
\caption{Accuracies on shortcut and anti-shortcut examples from RACE with different proportions of anti-shortcut examples in the training set, with the size set to 4k. The mean$\pm$standard deviations over 5 random seeds are displayed.}
\label{fig:blend-anti-mcqa}
\end{figure*}

\paragraph{Setup}
We used the online code \cite{rissanen1984universal} to approximate MDL.
In this algorithm, a training set is given to a model in a sequence of portions.
At each step, a model is trained from scratch on the portions given up to that point and is used to predict the next portion.
Practically, when the dataset is split into $S$ subsets with the time steps set to $\{t_1, t_2, ..., t_S\}$\footnote{The time steps were 0.1, 0.2, 0.4, 0.8, 1.6, 3.2, 6.25, 12.5, 25, 50, and 100 percent of the datasets following \citet{voita-titov-2020-information}.}, the MDL is estimated with the online code as follows:
\begin{align}
L = \sum_{i=0}^{S-1} \sum_{n=t_i+1}^{t_{i+1}} - \log_2 p_{\theta_i}(y_n|x_n),
\label{eq:mdl}
\end{align}
where $\theta_i$ is the parameter of a QA model trained on $\{(x_j, y_j)\}_{j=1}^{t_i}$ and $p_{\theta_0}$ is the uniform distribution.
Intuitively, the online code is related to the area under the loss curve and measures how much effort is required for the training.
See \citet{voita-titov-2020-information,pmlr-v139-perez21a} for more details about the online code.
The sizes of the biased dataset were 1400, 4000, 3000, and 300 for SQuAD 1.1, NaturalQuestions, RACE, and ReClor, respectively.
The size was set equally for each shortcut within a dataset.

\paragraph{Results}
The results are shown in Table \ref{tab:mdl}.
Note that the MDLs cannot be compared across datasets because the MDLs are dependent on the dataset size $t_S$ as shown in Eq. \ref{eq:mdl}.
For SQuAD 1.1 and NaturalQuestions, the availability of the answer-position shortcut made the dataset the easiest to learn among the three shortcuts, with the exception of RoBERTa on SQuAD 1.1.
The exception may be because RoBERTa can learn the word matching shortcut better than BERT as shown in Figure \ref{fig:training-dynamics}.
The MDLs for the word and type matching shortcuts differed for SQuAD 1.1 and NaturalQuestions.
For RACE and ReClor, the availability of the word-label correlation shortcut achieved lower MDLs than that of the lexical overlap shortcut.
Except for some cases, these observations align with the results of our behavioral tests in Figure \ref{fig:training-dynamics} and visualization in Figures \ref{fig:landscape-exqa} and \ref{fig:landscape-mcqa}.

In addition, RoBERTa consistently lowered the MDLs compared to BERT in all the cases.
Given that RoBERTa was more robust to anti-shortcut examples than BERT in Figure \ref{fig:training-dynamics}, the MDLs may also reflect the generalization capability of models as well as the characteristics of shortcuts.

\subsection{Balancing Shortcut and Anti-shortcut Examples}

\paragraph{RQ4} What proportion of anti-shortcut examples in a training set is required to avoid learning a shortcut? Is it related to the learnability of shortcuts?

One of the simplest approaches to mitigate shortcut learning is to reduce the dataset bias by adding anti-shortcut examples to training sets manually or automatically.
When a training set contains unintended biases or annotation artifacts, and the majority is solvable with shortcut solutions, models that adopt the shortcuts achieve low loss on the training set.
Therefore, increasing the proportion of anti-shortcut examples is a promising approach to avoid learning shortcuts \cite{lovering2021predicting}.

In addition, \citet{lovering2021predicting} showed that the requirement of the proportion of anti-shortcut examples is related to the extractability of shortcut cues.
We assume that there should be a similar relationship in QA datasets.
If we know how many anti-shortcut examples are required to avoid learning shortcuts, the knowledge can be utilized to construct new QA training sets or design data augmentation approaches \cite{yang-etal-2017-semi,shinoda-etal-2021-question} to make QA models learn more generalizable solutions.

\paragraph{Setup}
We changed the proportion of anti-shortcut examples from 0 to 1 with the sizes of the training sets fixed as 5k and 4k for extractive and multiple-choice QA, respectively.
For example, for the answer-position shortcut, the proportion of $\overline{\mathcal{D}_{\rm Position}}$ was changed from 0 to 1, and the scores on $\mathcal{D}_{\rm Position}$ and $\overline{\mathcal{D}_{\rm Position}}$ were reported.
We conducted the experiment for each shortcut separately on SQuAD 1.1 and RACE using BERT-base.

\paragraph{Results}
Figures \ref{fig:blend-anti-exqa} and \ref{fig:blend-anti-mcqa} show the results.
When the training sets consist of only shortcut examples, i.e., the x-axis value is 0, the gaps between the scores on $\mathcal{D}_{\rm k}$ and $\overline{\mathcal{D}_{\rm k}}$ are significant for all the cases.
When the proportion of anti-shortcut examples is 0.7, 0.8, and 0.9, the scores on $\mathcal{D}_{\rm k}$ and $\overline{\mathcal{D}_{\rm k}}$ are equal for Position, Top-1, and Overlap, respectively.
At these points, models do not use the shortcut but a solution that is equally generalizable to both the subsets.
In contrast, increasing the proportion of anti-shortcut examples more than these points degraded the scores on $\mathcal{D}_{\rm k}$.

When considering the learnability of each shortcut studied in our previous experiments, it is clear that more learnable shortcuts require a smaller proportion of anti-shortcut examples to achieve comparable performance on shortcut and anti-shortcut examples.
Moreover, for less-learnable shortcuts, such as Word and Type, we find that the score on $\mathcal{D}_{\rm k}$ is greater than that on $\overline{\mathcal{D}_{\rm k}}$ for almost all the points.
The results suggests that controlling the proportion of anti-shortcut examples alone is insufficient to mitigate the learning of less-learnable shortcuts.
For these less-learnable shortcuts, we may need to apply model-centric approaches such as \citet{clark-etal-2019-dont} to further mitigate the gap.

\section{Related Work}
Shortcut learning in deep neural networks (DNNs) \cite{geirhos_shortcut_2020} has received significant interests because it degrades the generalization of DNNs, causing humans to lose trust in AI \cite{10.1145/3442188.3445923}.
QA models for reading comprehension are no exception.
Although QA models have achieved human-level performance on some benchmarks \cite{rajpurkar-etal-2016-squad}, they lack robustness to challenging test sets such as adversarial attacks \cite{jia-liang-2017-adversarial}, questions that cannot be solved with partial-input baselines \cite{sugawara-etal-2018-makes}, paraphrased questions \cite{gan-ng-2019-improving}, answers in unseen positions \cite{ko-etal-2020-look}, and natural perturbations \cite{gardner-etal-2020-evaluating}.

The causes of this problem can be grouped into two categories: dataset and model.
For the data-centric cause, existing studies have found that substantial amounts of examples in QA datasets are solvable with question-answer type matching \cite{weissenborn-etal-2017-making} and word matching \cite{sugawara-etal-2018-makes} for extractive QA, and partial-input baselines \cite{sugawara-etal-2020-assessing,yu2020reclor} for multiple-choice QA.
As such, various shortcut solutions in QA have been studied individually.
To counter these problems, data augmentation approaches have been studied in QA.
\citet{jiang-bansal-2019-avoiding} constructed adversarial documents.
\citet{bartolo-etal-2020-beat} proposed model-in-the-loop annotation.
\citet{shinoda-etal-2021-improving} found that automatic question-answer pair generation can improve the robustness.

For the model-centric cause, several approaches have been applied to QA.
\citet{ko-etal-2020-look} used ensemble-based methods to unlearn an answer-position shortcut.
\citet{wu-etal-2020-improving} proposed concurrent modeling of multiple biases.
\citet{liu2020adversarial} used virtual adversarial training to improve the robustness to adversarial attacks.
\citet{wang2021infobert} introduced mutual-information-based regularizers.

In contrast to the above studies, several studies have attempted to understand shortcut learning.
\citet{lai-etal-2021-machine} found that shortcut solutions are learned at the early stage of training compared to a sophisticated solution on SQuAD.
\citet{lovering2021predicting} showed that the more extractable a shortcut cue with a probing classifier, the more anti-shortcut examples are needed to achieve low error on anti-shortcut examples in simple grammatical tasks.
\citet{scimeca2022which} compared several shortcut cues in image classification tasks.

We also attempt to understand the characteristics of shortcuts in extractive and multiple-choice QA from the perspectives of the learnability, that is, how easy it is to learn a shortcut.
To the best of our knowledge, we are the first to compare the difference of the learnability for each shortcut in QA.
Moreover, our study suggests that the learnability of shortcuts should be considered when designing mitigation methods.
This perspective is lacking in the existing mitigation studies.

\section{Conclusion}
We deepened understanding of the shortcut solutions in extractive and multiple-choice QA by comparing the learnability of shortcuts, that is, how easy it is to learn a shortcut, in a series of experiments.
We first showed that when every shortcut is applicable to a training set, extractive QA models prefer the answer-position shortcut whereas multiple-choice QA models prefer the word-label correlation shortcut among the examined shortcuts.
From the perspective of the parameter space, QA models that learn the preferred shortcuts tend to lie in flatter and deeper loss surfaces, which explains the cause of the preference.
To quantify the learnability of each shortcut, we estimated the MDLs on biased datasets where only one shortcut is valid.
The experimental results showed that the availability of more preferred shortcuts tend to make the task easier to learn.
To mitigate the shortcut learning behavior, we showed that more learnable shortcuts require less proportion of anti-shortcut examples during training.
The results also suggested that controlling the proportion of anti-shortcut examples alone is insufficient to avoid learning less-learnable shortcuts such as word and type matching in extractive QA.
We claim that approaches for mitigating shortcut learning should be appropriately designed according to the learnability of shortcuts.

\section*{Acknowledgements}
We would like to thank the anonymous reviewers for their valuable comments.
This work was supported by JSPS KAKENHI Grant Numbers 21H03502, 22J13751 and 22K17954.
This work was also supported by NEDO SIP-2 ``Big-data and AI-enabled Cyberspace Technologies''.

\bibliography{aaai23,anthology}

\begin{thebibliography}{41}
\providecommand{\natexlab}[1]{#1}

\bibitem[{Bartolo et~al.(2020)Bartolo, Roberts, Welbl, Riedel, and
  Stenetorp}]{bartolo-etal-2020-beat}
Bartolo, M.; Roberts, A.; Welbl, J.; Riedel, S.; and Stenetorp, P. 2020.
\newblock Beat the {AI}: Investigating Adversarial Human Annotation for Reading
  Comprehension.
\newblock \emph{Transactions of the Association for Computational Linguistics},
  8: 662--678.

\bibitem[{Clark, Yatskar, and Zettlemoyer(2019)}]{clark-etal-2019-dont}
Clark, C.; Yatskar, M.; and Zettlemoyer, L. 2019.
\newblock Don{'}t Take the Easy Way Out: Ensemble Based Methods for Avoiding
  Known Dataset Biases.
\newblock In \emph{Proceedings of the 2019 Conference on Empirical Methods in
  Natural Language Processing and the 9th International Joint Conference on
  Natural Language Processing (EMNLP-IJCNLP)}, 4069--4082. Hong Kong, China:
  Association for Computational Linguistics.

\bibitem[{Devlin et~al.(2019)Devlin, Chang, Lee, and
  Toutanova}]{devlin-etal-2019-bert}
Devlin, J.; Chang, M.-W.; Lee, K.; and Toutanova, K. 2019.
\newblock {BERT}: Pre-training of Deep Bidirectional Transformers for Language
  Understanding.
\newblock In \emph{Proceedings of the 2019 Conference of the North {A}merican
  Chapter of the Association for Computational Linguistics: Human Language
  Technologies, Volume 1 (Long and Short Papers)}, 4171--4186. Minneapolis,
  Minnesota: Association for Computational Linguistics.

\bibitem[{Gan and Ng(2019)}]{gan-ng-2019-improving}
Gan, W.~C.; and Ng, H.~T. 2019.
\newblock Improving the Robustness of Question Answering Systems to Question
  Paraphrasing.
\newblock In \emph{Proceedings of the 57th Annual Meeting of the Association
  for Computational Linguistics}, 6065--6075. Florence, Italy: Association for
  Computational Linguistics.

\bibitem[{Gardner et~al.(2020)Gardner, Artzi, Basmov, Berant, Bogin, Chen,
  Dasigi, Dua, Elazar, Gottumukkala, Gupta, Hajishirzi, Ilharco, Khashabi, Lin,
  Liu, Liu, Mulcaire, Ning, Singh, Smith, Subramanian, Tsarfaty, Wallace,
  Zhang, and Zhou}]{gardner-etal-2020-evaluating}
Gardner, M.; Artzi, Y.; Basmov, V.; Berant, J.; Bogin, B.; Chen, S.; Dasigi,
  P.; Dua, D.; Elazar, Y.; Gottumukkala, A.; Gupta, N.; Hajishirzi, H.;
  Ilharco, G.; Khashabi, D.; Lin, K.; Liu, J.; Liu, N.~F.; Mulcaire, P.; Ning,
  Q.; Singh, S.; Smith, N.~A.; Subramanian, S.; Tsarfaty, R.; Wallace, E.;
  Zhang, A.; and Zhou, B. 2020.
\newblock Evaluating Models{'} Local Decision Boundaries via Contrast Sets.
\newblock In \emph{Findings of the Association for Computational Linguistics:
  EMNLP 2020}, 1307--1323. Online: Association for Computational Linguistics.

\bibitem[{Gardner et~al.(2021)Gardner, Merrill, Dodge, Peters, Ross, Singh, and
  Smith}]{gardner-etal-2021-competency}
Gardner, M.; Merrill, W.; Dodge, J.; Peters, M.; Ross, A.; Singh, S.; and
  Smith, N.~A. 2021.
\newblock Competency Problems: On Finding and Removing Artifacts in Language
  Data.
\newblock In \emph{Proceedings of the 2021 Conference on Empirical Methods in
  Natural Language Processing}, 1801--1813. Online and Punta Cana, Dominican
  Republic: Association for Computational Linguistics.

\bibitem[{Geirhos et~al.(2020)Geirhos, Jacobsen, Michaelis, Zemel, Brendel,
  Bethge, and Wichmann}]{geirhos_shortcut_2020}
Geirhos, R.; Jacobsen, J.-H.; Michaelis, C.; Zemel, R.; Brendel, W.; Bethge,
  M.; and Wichmann, F.~A. 2020.
\newblock Shortcut learning in deep neural networks.
\newblock \emph{Nature Machine Intelligence}, 2(11): 665--673.

\bibitem[{Gururangan et~al.(2018)Gururangan, Swayamdipta, Levy, Schwartz,
  Bowman, and Smith}]{gururangan-etal-2018-annotation}
Gururangan, S.; Swayamdipta, S.; Levy, O.; Schwartz, R.; Bowman, S.; and Smith,
  N.~A. 2018.
\newblock Annotation Artifacts in Natural Language Inference Data.
\newblock In \emph{Proceedings of the 2018 Conference of the North {A}merican
  Chapter of the Association for Computational Linguistics: Human Language
  Technologies, Volume 2 (Short Papers)}, 107--112. New Orleans, Louisiana:
  Association for Computational Linguistics.

\bibitem[{Hochreiter and Schmidhuber(1997)}]{10.1162/neco.1997.9.1.1}
Hochreiter, S.; and Schmidhuber, J. 1997.
\newblock {Flat Minima}.
\newblock \emph{Neural Computation}, 9(1): 1--42.

\bibitem[{Honnibal et~al.(2020)Honnibal, Montani, Van~Landeghem, and
  Boyd}]{spacy2}
Honnibal, M.; Montani, I.; Van~Landeghem, S.; and Boyd, A. 2020.
\newblock {spaCy: Industrial-strength Natural Language Processing in Python}.

\bibitem[{Hu et~al.(2020)Hu, Xiao, Adlam, and
  Pennington}]{10.5555/3495724.3497160}
Hu, W.; Xiao, L.; Adlam, B.; and Pennington, J. 2020.
\newblock The Surprising Simplicity of the Early-Time Learning Dynamics of
  Neural Networks.
\newblock In \emph{Proceedings of the 34th International Conference on Neural
  Information Processing Systems}, NIPS'20. Red Hook, NY, USA: Curran
  Associates Inc.
\newblock ISBN 9781713829546.

\bibitem[{Jacovi et~al.(2021)Jacovi, Marasovi\'{c}, Miller, and
  Goldberg}]{10.1145/3442188.3445923}
Jacovi, A.; Marasovi\'{c}, A.; Miller, T.; and Goldberg, Y. 2021.
\newblock Formalizing Trust in Artificial Intelligence: Prerequisites, Causes
  and Goals of Human Trust in AI.
\newblock In \emph{Proceedings of the 2021 ACM Conference on Fairness,
  Accountability, and Transparency}, FAccT '21, 624–635. New York, NY, USA:
  Association for Computing Machinery.
\newblock ISBN 9781450383097.

\bibitem[{Jia and Liang(2017)}]{jia-liang-2017-adversarial}
Jia, R.; and Liang, P. 2017.
\newblock Adversarial Examples for Evaluating Reading Comprehension Systems.
\newblock In \emph{Proceedings of the 2017 Conference on Empirical Methods in
  Natural Language Processing}, 2021--2031. Copenhagen, Denmark: Association
  for Computational Linguistics.

\bibitem[{Jiang and Bansal(2019)}]{jiang-bansal-2019-avoiding}
Jiang, Y.; and Bansal, M. 2019.
\newblock Avoiding Reasoning Shortcuts: Adversarial Evaluation, Training, and
  Model Development for Multi-Hop {QA}.
\newblock In \emph{Proceedings of the 57th Annual Meeting of the Association
  for Computational Linguistics}, 2726--2736. Florence, Italy: Association for
  Computational Linguistics.

\bibitem[{Ko et~al.(2020)Ko, Lee, Kim, Kim, and Kang}]{ko-etal-2020-look}
Ko, M.; Lee, J.; Kim, H.; Kim, G.; and Kang, J. 2020.
\newblock Look at the First Sentence: Position Bias in Question Answering.
\newblock In \emph{Proceedings of the 2020 Conference on Empirical Methods in
  Natural Language Processing (EMNLP)}, 1109--1121. Online: Association for
  Computational Linguistics.

\bibitem[{Kwiatkowski et~al.(2019)Kwiatkowski, Palomaki, Redfield, Collins,
  Parikh, Alberti, Epstein, Polosukhin, Devlin, Lee, Toutanova, Jones, Kelcey,
  Chang, Dai, Uszkoreit, Le, and Petrov}]{kwiatkowski-etal-2019-natural}
Kwiatkowski, T.; Palomaki, J.; Redfield, O.; Collins, M.; Parikh, A.; Alberti,
  C.; Epstein, D.; Polosukhin, I.; Devlin, J.; Lee, K.; Toutanova, K.; Jones,
  L.; Kelcey, M.; Chang, M.-W.; Dai, A.~M.; Uszkoreit, J.; Le, Q.; and Petrov,
  S. 2019.
\newblock Natural Questions: A Benchmark for Question Answering Research.
\newblock \emph{Transactions of the Association for Computational Linguistics},
  7: 452--466.

\bibitem[{Lai et~al.(2017)Lai, Xie, Liu, Yang, and Hovy}]{lai-etal-2017-race}
Lai, G.; Xie, Q.; Liu, H.; Yang, Y.; and Hovy, E. 2017.
\newblock {RACE}: Large-scale {R}e{A}ding Comprehension Dataset From
  Examinations.
\newblock In \emph{Proceedings of the 2017 Conference on Empirical Methods in
  Natural Language Processing}, 785--794. Copenhagen, Denmark: Association for
  Computational Linguistics.

\bibitem[{Lai et~al.(2021)Lai, Zhang, Feng, Huang, and
  Zhao}]{lai-etal-2021-machine}
Lai, Y.; Zhang, C.; Feng, Y.; Huang, Q.; and Zhao, D. 2021.
\newblock Why Machine Reading Comprehension Models Learn Shortcuts?
\newblock In \emph{Findings of the Association for Computational Linguistics:
  ACL-IJCNLP 2021}, 989--1002. Online: Association for Computational
  Linguistics.

\bibitem[{Li et~al.(2018)Li, Xu, Taylor, Studer, and
  Goldstein}]{NEURIPS2018_a41b3bb3}
Li, H.; Xu, Z.; Taylor, G.; Studer, C.; and Goldstein, T. 2018.
\newblock Visualizing the Loss Landscape of Neural Nets.
\newblock In Bengio, S.; Wallach, H.; Larochelle, H.; Grauman, K.;
  Cesa-Bianchi, N.; and Garnett, R., eds., \emph{Advances in Neural Information
  Processing Systems}, volume~31. Curran Associates, Inc.

\bibitem[{Liu et~al.(2020)Liu, Cheng, He, Chen, Wang, Poon, and
  Gao}]{liu2020adversarial}
Liu, X.; Cheng, H.; He, P.; Chen, W.; Wang, Y.; Poon, H.; and Gao, J. 2020.
\newblock Adversarial training for large neural language models.
\newblock \emph{arXiv preprint arXiv:2004.08994}.

\bibitem[{Liu et~al.(2019)Liu, Ott, Goyal, Du, Joshi, Chen, Levy, Lewis,
  Zettlemoyer, and Stoyanov}]{liu2019roberta}
Liu, Y.; Ott, M.; Goyal, N.; Du, J.; Joshi, M.; Chen, D.; Levy, O.; Lewis, M.;
  Zettlemoyer, L.; and Stoyanov, V. 2019.
\newblock Roberta: A robustly optimized bert pretraining approach.
\newblock \emph{arXiv preprint arXiv:1907.11692}.

\bibitem[{Lovering et~al.(2021)Lovering, Jha, Linzen, and
  Pavlick}]{lovering2021predicting}
Lovering, C.; Jha, R.; Linzen, T.; and Pavlick, E. 2021.
\newblock Predicting Inductive Biases of Pre-Trained Models.
\newblock In \emph{International Conference on Learning Representations}.

\bibitem[{McCoy, Pavlick, and Linzen(2019)}]{mccoy-etal-2019-right}
McCoy, T.; Pavlick, E.; and Linzen, T. 2019.
\newblock Right for the Wrong Reasons: Diagnosing Syntactic Heuristics in
  Natural Language Inference.
\newblock In \emph{Proceedings of the 57th Annual Meeting of the Association
  for Computational Linguistics}, 3428--3448. Florence, Italy: Association for
  Computational Linguistics.

\bibitem[{Perez, Kiela, and Cho(2021)}]{pmlr-v139-perez21a}
Perez, E.; Kiela, D.; and Cho, K. 2021.
\newblock Rissanen Data Analysis: Examining Dataset Characteristics via
  Description Length.
\newblock In Meila, M.; and Zhang, T., eds., \emph{Proceedings of the 38th
  International Conference on Machine Learning}, volume 139 of
  \emph{Proceedings of Machine Learning Research}, 8500--8513. PMLR.

\bibitem[{Rajpurkar et~al.(2016)Rajpurkar, Zhang, Lopyrev, and
  Liang}]{rajpurkar-etal-2016-squad}
Rajpurkar, P.; Zhang, J.; Lopyrev, K.; and Liang, P. 2016.
\newblock {SQ}u{AD}: 100,000+ Questions for Machine Comprehension of Text.
\newblock In \emph{Proceedings of the 2016 Conference on Empirical Methods in
  Natural Language Processing}, 2383--2392. Austin, Texas: Association for
  Computational Linguistics.

\bibitem[{Rissanen(1978)}]{RISSANEN1978465}
Rissanen, J. 1978.
\newblock Modeling by shortest data description.
\newblock \emph{Automatica}, 14(5): 465--471.

\bibitem[{Rissanen(1984)}]{rissanen1984universal}
Rissanen, J. 1984.
\newblock Universal coding, information, prediction, and estimation.
\newblock \emph{IEEE Transactions on Information theory}, 30(4): 629--636.

\bibitem[{Scimeca et~al.(2022)Scimeca, Oh, Chun, Poli, and
  Yun}]{scimeca2022which}
Scimeca, L.; Oh, S.~J.; Chun, S.; Poli, M.; and Yun, S. 2022.
\newblock Which Shortcut Cues Will {DNN}s Choose? A Study from the
  Parameter-Space Perspective.
\newblock In \emph{International Conference on Learning Representations}.

\bibitem[{Shinoda, Sugawara, and
  Aizawa(2021{\natexlab{a}})}]{shinoda-etal-2021-question}
Shinoda, K.; Sugawara, S.; and Aizawa, A. 2021{\natexlab{a}}.
\newblock Can Question Generation Debias Question Answering Models? A Case
  Study on Question{--}Context Lexical Overlap.
\newblock In \emph{Proceedings of the 3rd Workshop on Machine Reading for
  Question Answering}, 63--72. Punta Cana, Dominican Republic: Association for
  Computational Linguistics.

\bibitem[{Shinoda, Sugawara, and
  Aizawa(2021{\natexlab{b}})}]{shinoda-etal-2021-improving}
Shinoda, K.; Sugawara, S.; and Aizawa, A. 2021{\natexlab{b}}.
\newblock Improving the Robustness of {QA} Models to Challenge Sets with
  Variational Question-Answer Pair Generation.
\newblock In \emph{Proceedings of the 59th Annual Meeting of the Association
  for Computational Linguistics and the 11th International Joint Conference on
  Natural Language Processing: Student Research Workshop}, 197--214. Online:
  Association for Computational Linguistics.

\bibitem[{Sugawara et~al.(2018)Sugawara, Inui, Sekine, and
  Aizawa}]{sugawara-etal-2018-makes}
Sugawara, S.; Inui, K.; Sekine, S.; and Aizawa, A. 2018.
\newblock What Makes Reading Comprehension Questions Easier?
\newblock In \emph{Proceedings of the 2018 Conference on Empirical Methods in
  Natural Language Processing}, 4208--4219. Brussels, Belgium: Association for
  Computational Linguistics.

\bibitem[{Sugawara et~al.(2020)Sugawara, Stenetorp, Inui, and
  Aizawa}]{sugawara-etal-2020-assessing}
Sugawara, S.; Stenetorp, P.; Inui, K.; and Aizawa, A. 2020.
\newblock Assessing the {Benchmarking} {Capacity} of {Machine} {Reading}
  {Comprehension} {Datasets}.
\newblock \emph{Proceedings of the AAAI Conference on Artificial Intelligence},
  34(05): 8918--8927.

\bibitem[{Torralba and Efros(2011)}]{5995347}
Torralba, A.; and Efros, A.~A. 2011.
\newblock Unbiased look at dataset bias.
\newblock In \emph{CVPR 2011}, 1521--1528.

\bibitem[{Utama, Moosavi, and Gurevych(2020)}]{utama-etal-2020-towards}
Utama, P.~A.; Moosavi, N.~S.; and Gurevych, I. 2020.
\newblock Towards Debiasing {NLU} Models from Unknown Biases.
\newblock In \emph{Proceedings of the 2020 Conference on Empirical Methods in
  Natural Language Processing (EMNLP)}, 7597--7610. Online: Association for
  Computational Linguistics.

\bibitem[{Vaswani et~al.(2017)Vaswani, Shazeer, Parmar, Uszkoreit, Jones,
  Gomez, Kaiser, and Polosukhin}]{NIPS2017_3f5ee243}
Vaswani, A.; Shazeer, N.; Parmar, N.; Uszkoreit, J.; Jones, L.; Gomez, A.~N.;
  Kaiser, L.~u.; and Polosukhin, I. 2017.
\newblock Attention is All you Need.
\newblock In Guyon, I.; Luxburg, U.~V.; Bengio, S.; Wallach, H.; Fergus, R.;
  Vishwanathan, S.; and Garnett, R., eds., \emph{Advances in Neural Information
  Processing Systems}, volume~30. Curran Associates, Inc.

\bibitem[{Voita and Titov(2020)}]{voita-titov-2020-information}
Voita, E.; and Titov, I. 2020.
\newblock Information-Theoretic Probing with Minimum Description Length.
\newblock In \emph{Proceedings of the 2020 Conference on Empirical Methods in
  Natural Language Processing (EMNLP)}, 183--196. Online: Association for
  Computational Linguistics.

\bibitem[{Wang et~al.(2021)Wang, Wang, Cheng, Gan, Jia, Li, and
  Liu}]{wang2021infobert}
Wang, B.; Wang, S.; Cheng, Y.; Gan, Z.; Jia, R.; Li, B.; and Liu, J. 2021.
\newblock Info{\{}BERT{\}}: Improving Robustness of Language Models from An
  Information Theoretic Perspective.
\newblock In \emph{International Conference on Learning Representations}.

\bibitem[{Weissenborn, Wiese, and Seiffe(2017)}]{weissenborn-etal-2017-making}
Weissenborn, D.; Wiese, G.; and Seiffe, L. 2017.
\newblock Making Neural {QA} as Simple as Possible but not Simpler.
\newblock In \emph{Proceedings of the 21st Conference on Computational Natural
  Language Learning ({C}o{NLL} 2017)}, 271--280. Vancouver, Canada: Association
  for Computational Linguistics.

\bibitem[{Wu et~al.(2020)Wu, Moosavi, R{\"u}ckl{\'e}, and
  Gurevych}]{wu-etal-2020-improving}
Wu, M.; Moosavi, N.~S.; R{\"u}ckl{\'e}, A.; and Gurevych, I. 2020.
\newblock Improving {QA} Generalization by Concurrent Modeling of Multiple
  Biases.
\newblock In \emph{Findings of the Association for Computational Linguistics:
  EMNLP 2020}, 839--853. Online: Association for Computational Linguistics.

\bibitem[{Yang et~al.(2017)Yang, Hu, Salakhutdinov, and
  Cohen}]{yang-etal-2017-semi}
Yang, Z.; Hu, J.; Salakhutdinov, R.; and Cohen, W. 2017.
\newblock Semi-Supervised {QA} with Generative Domain-Adaptive Nets.
\newblock In \emph{Proceedings of the 55th Annual Meeting of the Association
  for Computational Linguistics (Volume 1: Long Papers)}, 1040--1050.
  Vancouver, Canada: Association for Computational Linguistics.

\bibitem[{Yu et~al.(2020)Yu, Jiang, Dong, and Feng}]{yu2020reclor}
Yu, W.; Jiang, Z.; Dong, Y.; and Feng, J. 2020.
\newblock ReClor: A Reading Comprehension Dataset Requiring Logical Reasoning.
\newblock In \emph{International Conference on Learning Representations
  (ICLR)}.

\end{thebibliography}
\end{document}